\documentclass{article} 
\usepackage{arxiv}
\usepackage{graphicx}
\usepackage{multirow}
\usepackage{multicol}
\usepackage{algorithm}
\usepackage{algorithmic}
\usepackage{amsmath}
\usepackage{amsfonts}
\usepackage{bm}
\usepackage{enumitem}
\usepackage{subfig} 
\usepackage{booktabs}

\usepackage{listings} 
\usepackage{xcolor} 
\usepackage{url}

\usepackage[
    backend=biber,
    citestyle=authoryear,
    bibstyle=authortitle,
    mincitenames=1,
    maxcitenames=1,
    maxbibnames=3,
]{biblatex}
\def\Figref#1{Figure~\ref{#1}}

\def\Secref#1{Section~\ref{#1}}

\def\eqref#1{equation~\ref{#1}}
\def\Eqref#1{Equation~\ref{#1}}

\def\1{\bm{1}}

\def\vq{{\bm{q}}}

\def\mG{{\bm{G}}}

\def\mK{{\bm{K}}}

\def\mV{{\bm{V}}}

\DeclareMathAlphabet{\mathsfit}{\encodingdefault}{\sfdefault}{m}{sl}
\SetMathAlphabet{\mathsfit}{bold}{\encodingdefault}{\sfdefault}{bx}{n}

\newcommand{\R}{\mathbb{R}}

\newcommand{\citep}[1]{\parencite{#1}}
\setlist[itemize,1]{leftmargin=\dimexpr 18pt}
\setlist[enumerate,1]{leftmargin=\dimexpr 18pt}

\title{
\raisebox{-0.1\height}{\includegraphics[width=0.032\textwidth]{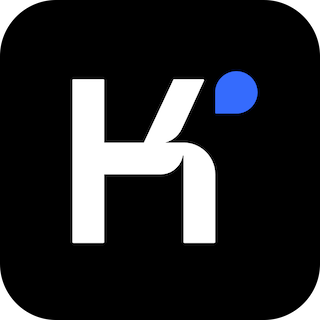}} %
MoBA: Mixture of Block Attention for Long-Context LLMs}

\renewcommand{\headeright}{Technical Report}
\renewcommand{\undertitle}{Technical Report}
\renewcommand{\shorttitle}{%
\raisebox{-0.1\height}{\includegraphics[width=0.020\textwidth]{fig/logo.png}} %
MoBA: Mixture of Block Attention
}

\author{%
    \textbf{Enzhe Lu}$^{1}$ \quad \textbf{Zhejun Jiang}$^{1}$ \quad \textbf{Jingyuan Liu}$^{1}$ \quad \textbf{Yulun Du}$^{1}$ \quad \textbf{Tao Jiang}$^{1}$ \quad \textbf{Chao Hong}$^{1}$ \\
    \textbf{Shaowei Liu}$^{1}$ \quad \textbf{Weiran He}$^{1}$ \quad \textbf{Enming Yuan}$^{1}$ \quad \textbf{Yuzhi Wang}$^{1}$ \quad \textbf{Zhiqi Huang}$^{1}$ \quad \textbf{Huan Yuan}$^{1}$  \\
    \textbf{Suting Xu}$^{1}$ \quad \textbf{Xinran Xu}$^{1}$ \quad \textbf{Guokun Lai}$^{1}$ \quad \textbf{Yanru Chen}$^{1}$ \quad \textbf{Huabin Zheng}$^{1}$ \quad \textbf{Junjie Yan}$^{1}$ \\
     \textbf{Jianlin Su}$^{1}$ \quad \textbf{Yuxin Wu}$^{1}$ \quad \textbf{Neo Y. Zhang}$^{1}$ \quad \textbf{Zhilin Yang}$^{1}$ \\
    \textbf{Xinyu Zhou}$^{1,\ddagger}$ \quad \textbf{Mingxing Zhang}$^{2,*}$ \quad \textbf{Jiezhong Qiu}$^{3,\ddagger}$ 
    \thanks{$^*$\texttt{zhang\_mingxing@mail.tsinghua.edu.cn}}
    \thanks{$^\ddagger$Co-corresponding authors. 
    Xinyu Zhou (\texttt{zhouxinyu@moonshot.cn}), 
    Jiezhong Qiu (\texttt{jiezhongqiu@outlook.com})}
    \\[2ex]
    $^1$ Moonshot AI \quad $^2$ Tsinghua University \quad $^3$ Zhejiang Lab/Zhejiang University
}

\date{}

\begin{document}

\maketitle

\begin{abstract}
Scaling the effective context length is essential for advancing large language models (LLMs) toward artificial general intelligence (AGI). However, the quadratic increase in computational complexity inherent in traditional attention mechanisms presents a prohibitive overhead. Existing approaches either impose strongly biased structures, such as sink or window attention which are task-specific, or radically modify the attention mechanism into linear approximations, whose performance in complex reasoning tasks remains inadequately explored.

In this work, we propose a solution that adheres to the ``less structure'' principle, allowing the model to determine where to attend autonomously, rather than introducing predefined biases. We introduce Mixture of Block Attention (MoBA), an innovative approach that applies the principles of Mixture of Experts (MoE) to the attention mechanism. This novel architecture demonstrates superior performance on long-context tasks while offering a key advantage: the ability to seamlessly transition between full and sparse attention, enhancing efficiency without the risk of compromising performance. MoBA has already been deployed to support Kimi's long-context requests and demonstrates significant advancements in efficient attention computation for LLMs. Our code is available at \url{https://github.com/MoonshotAI/MoBA}.
\end{abstract}

\section{Introduction}

The pursuit of artificial general intelligence (AGI) has driven the development of large language models (LLMs) to unprecedented scales, with the promise of handling complex tasks that mimic human cognition. A pivotal capability for achieving AGI is the ability to process, understand, and generate long sequences, which is essential for a wide range of applications, from historical data analysis to complex reasoning and decision-making processes.
This growing demand for extended context processing can be seen not only in the popularity of long input prompt understanding, as showcased by models like Kimi~\citep{kimi}, Claude~\citep{claude} and Gemini~\citep{reid2024gemini}, but also in recent explorations of long chain-of-thought (CoT) output capabilities in Kimi k1.5~\citep{team2025kimi}, DeepSeek-R1~\citep{guo2025deepseek}, and OpenAI o1/o3~\citep{guan2024deliberative}.

However, extending the sequence length in LLMs is non-trivial due to the quadratic growth in computational complexity associated with the vanilla attention mechanism~\citep{waswani2017attention}. This challenge has spurred a wave of research aimed at improving efficiency without sacrificing performance. One prominent direction capitalizes on the inherent sparsity of attention scores. This sparsity arises both mathematically --- from the softmax operation, where various sparse attention patterns have be studied~\citep{jiang2024minference} --- and biologically~\citep{watson2025human}, where sparse connectivity is observed in brain regions related to memory storage.

Existing approaches often leverage predefined structural constraints, such as sink-based~\citep{xiao2023efficient} or sliding window attention~\citep{beltagy2020longformer}, to exploit this sparsity. While these methods can be effective, they tend to be highly task-specific, potentially hindering the model’s overall generalizability. Alternatively, a range of dynamic sparse attention mechanisms, exemplified by Quest~\citep{tang2024quest}, Minference~\citep{jiang2024minference}, and RetrievalAttention~\citep{liu2024retrievalattention}, select subsets of tokens at inference time. Although such methods can reduce computation for long sequences, they do not substantially alleviate the intensive training costs of long-context models, making it challenging to scale LLMs efficiently to contexts on the order of millions of tokens.
Another promising alternative way has recently emerged in the form of linear attention models, such as Mamba~\citep{dao2024transformers}, RWKV~\citep{peng2023rwkv, peng2024eagle}, and RetNet~\citep{sun2023retentive}. These approaches replace canonical softmax-based attention with linear approximations, thereby reducing the computational overhead for long-sequence processing. However, due to the substantial differences between linear and conventional attention, adapting existing Transformer models typically incurs high conversion costs~\citep{mercat2024linearizing, wang2024mamba, bick2025transformers, zhang2024lolcats} or requires training entirely new models from scratch~\citep{li2025minimax}. More importantly, evidence of their effectiveness in complex reasoning tasks remains limited.

Consequently, a critical research question arises: How can we design a robust and adaptable attention architecture that retains the original Transformer framework while {\bf adhering to a ``less structure'' principle, allowing the model to determine where to attend without relying on predefined biases?} Ideally, such an architecture would transition seamlessly between full and sparse attention modes, thus maximizing compatibility with existing pre-trained models and enabling both efficient inference and accelerated training without compromising performance.

Thus, we introduce Mixture of Block Attention (MoBA), a novel architecture that builds upon the innovative principles of Mixture of Experts (MoE)~\citep{shazeer2017outrageously} and applies them to the attention mechanism of the Transformer model. MoE has been used primarily in the feedforward network~(FFN) layers of Transformers~\citep{lepikhin2020gshard,fedus2022switch, zoph2022st}, but MoBA pioneers its application to long context attention, allowing dynamic selection of historically relevant blocks of key and values for each query token. This approach not only enhances the efficiency of LLMs but also enables them to handle longer and more complex prompts without a proportional increase in resource consumption.
MoBA addresses the computational inefficiency of traditional attention mechanisms by partitioning the context into blocks and employing a gating mechanism to selectively route query tokens to the most relevant blocks. This block sparse attention significantly reduces the computational costs, paving the way for more efficient processing of long sequences. The model's ability to dynamically select the most informative blocks of keys leads to improved performance and efficiency, particularly beneficial for tasks involving extensive contextual information.

In this paper, we detail the architecture of MoBA, firstly its block partitioning and routing strategy, and secondly its computational efficiency compared to traditional attention mechanisms. We further present experimental results that demonstrate MoBA's superior performance in tasks requiring the processing of long sequences. Our work contributes a novel approach to efficient attention computation, pushing the boundaries of what is achievable with LLMs in handling complex and lengthy inputs.

\section{Method}
\label{sec:method}
\begin{figure}[t]
    \centering
    \subfloat[]{
        \includegraphics[width=0.45\textwidth]{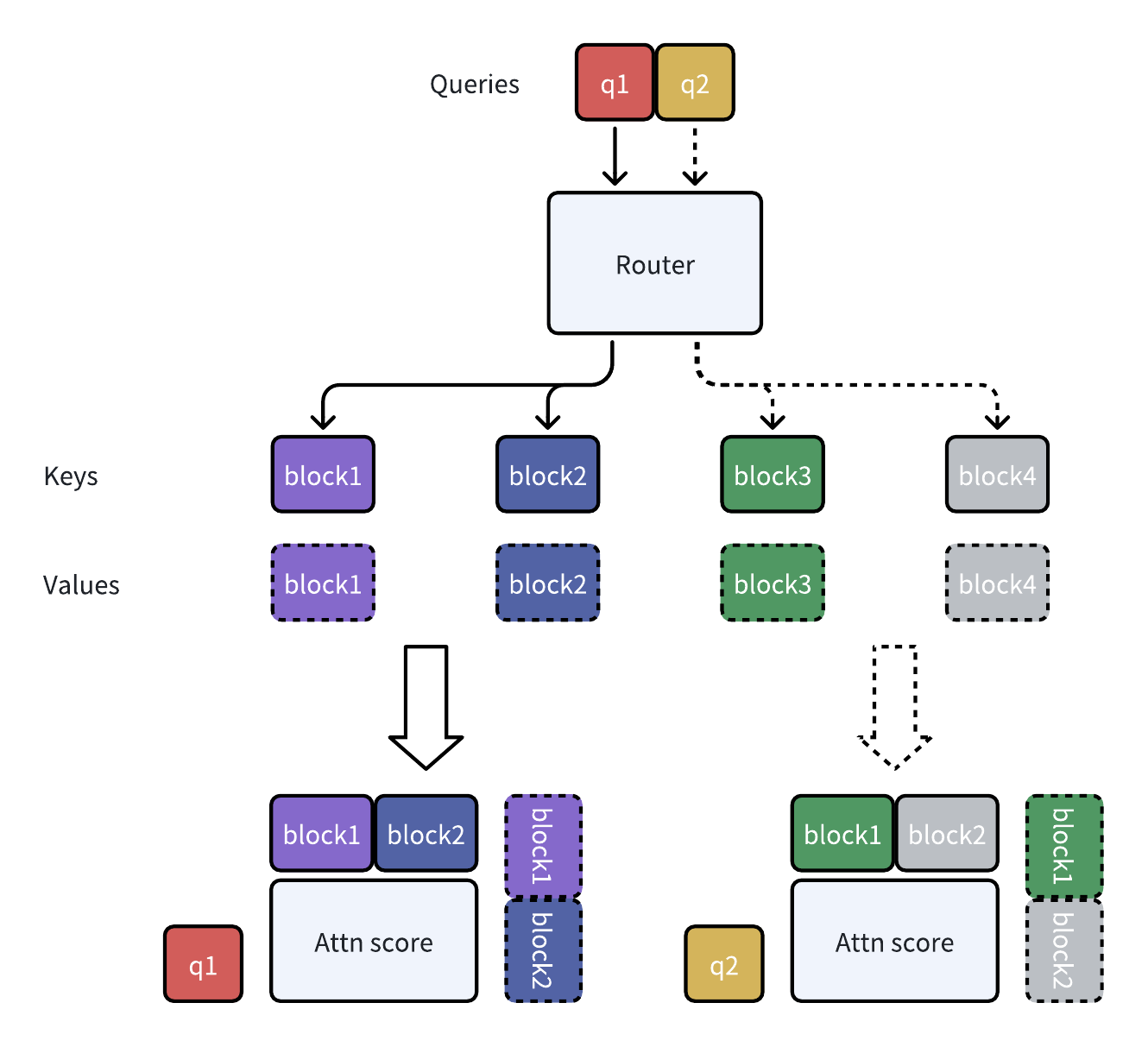}
        \label{fig:moba_example}
    }
    \subfloat[]{
        \includegraphics[width=0.4\textwidth]{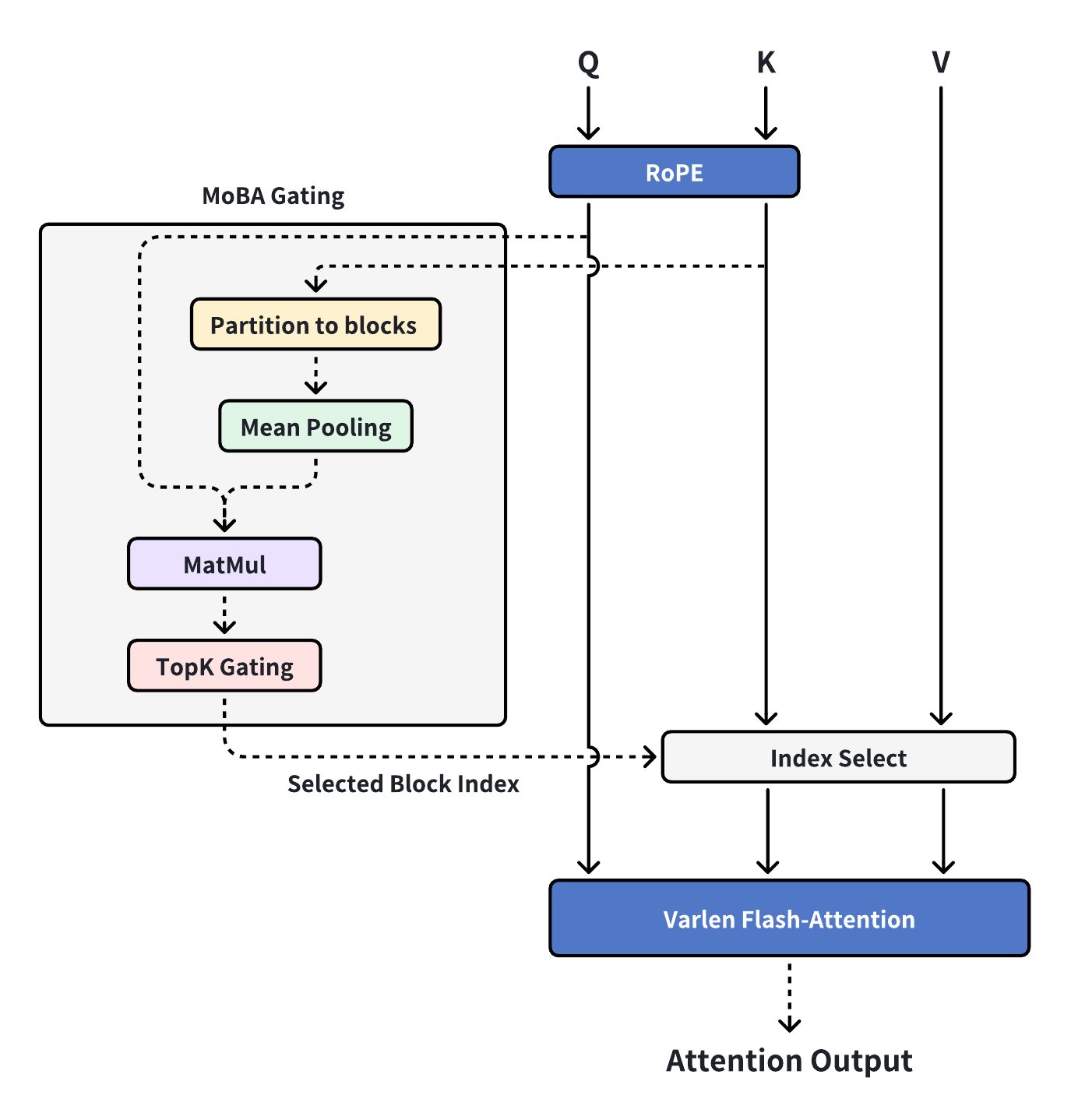}
        \label{fig:moba_flash}
    }
    \caption{Illustration of mixture of block attention~(MoBA).
    \textbf{(a)} A running example of MoBA; \textbf{(b)} Integration of MoBA into Flash Attention.
    } 
    \label{Fig:moba} 
\end{figure}

In this work, we introduce a novel architecture, termed Mixture of Block Attention (MoBA), which extends the capabilities of the Transformer model by dynamically selecting historical segments (blocks) for attention computation. MoBA is inspired by techniques of Mixture of Experts (MoE)  and sparse attention. The former technique has been predominantly applied to the feedforward network (FFN) layers within the Transformer architecture, while the latter has been widely adopted in scaling Transformers to handle long contexts. Our method is innovative in applying the MoE principle to the attention mechanism itself, allowing for more efficient and effective processing of long sequences.

\subsection{Preliminaries: Standard Attention in Transformer}

We first revisit the standard Attention in Transformers.
For simplicity, we revisit the case where a single query token $\vq \in \R^{1\times d}$
attends to the $N$ key and value tokens, denoting $\mK,\mV \in \R^{N\times d}$, respectively. The standard attention is computed as:
\begin{equation}
\mathrm{Attn}(\vq, \mK, \mV) = \mathrm{Softmax}{\left(\vq \mK^\top\right)}\mV,
\end{equation}
where $d$ denotes the dimension of a single attention head. We focus on the single-head scenario for clarity. The extension to multi-head attention involves concatenating the outputs from multiple such single-head attention operations.

\subsection{MoBA Architecture}

Different from standard attention where each query tokens attend to the entire context, MoBA enables each query token to only attend to a subset of keys and values:
\begin{equation}
\mathrm{MoBA}(\vq, \mK, \mV) = \mathrm{Softmax}{\left(\vq {\mK[I]}^\top\right)}\mV[I],
\end{equation}
where $I \subseteq [N]$ is the set of selected keys and values.

The key innovation in MoBA is the block partitioning and selection strategy. We divide the full context of length $N$ into $n$ blocks, where each block represents a subset of subsequent tokens. Without loss of generality, we assume that the context length $N$ is divisible by the number of blocks $n$. We  further denote  $B =\frac{N}{n}$ to be the block size and
\begin{equation}
I_i = \left[(i-1)\times B+1, i \times B\right]
\end{equation}to be the range of the $i$-th block.
By applying the top-$k$ gating mechanism from MoE, we enable each query to selectively focus on a subset of tokens from different blocks, rather than the entire context:
\begin{equation}
\label{eq:moba_gate}
I = \bigcup_{g_i > 0} I_i. 
\end{equation}
The model employs a gating mechanism, as $g_i$ in \Eqref{eq:moba_gate}, to select the most relevant blocks for each query token. 
The MoBA gate first computes the affinity score $s_i$ measuring the relevance between query $\vq$ and the $i$-th block, and applies a top-$k$ gating among all blocks. More formally, the gate value for the $i$-th block $g_i$ is computed by
\begin{equation}
    g_i = \begin{cases} 
1 & s_i \in \mathrm{Topk}\left(\{s_j | j\in [n]\}, k\right) \\
0 & \text{otherwise}\end{cases},
\end{equation}where $\mathrm{Topk}(\cdot, k)$ denotes the set containing $k$ highest scores among the affinity scores calculated for each block.
In this work, the score $s_i$ is computed by the inner product between $\vq$ and the mean pooling of $\mK[I_i]$ along the sequence dimension:
\begin{equation}
\label{eq:gating_score}
    s_i = \langle \vq, \mathrm{mean\_pool}(\mK[I_i])\rangle
\end{equation}

\textbf{A Running Example.} We provide a running example of MoBA at \Figref{fig:moba_example}, where we have two query tokens and four KV blocks. The router~(gating network) dynamically selects the top two blocks for each query to attend. As shown in \Figref{fig:moba_example}, the first query is assigned to the first and second blocks, while the second query is assigned to the third and fourth blocks.

It is important to maintain causality in autoregressive language models, as they generate text by next-token prediction based on previous tokens. This sequential generation process ensures that a token cannot influence tokens that come before it, thus preserving the causal relationship.
MoBA preserves causality through two specific designs: 

\textbf{Causality: No Attention to Future Blocks.} MoBA ensures that a query token cannot be routed to any future blocks. By limiting the attention scope to current and past blocks, MoBA adheres to the autoregressive nature of language modeling. More formally, denoting $\mathrm{pos}(\vq)$ as the position index of the query $\vq$, we set $s_i = -\infty$ and $g_i = 0$  for any blocks $i$ such that $\mathrm{pos}(\vq) < i \times B$.

\textbf{Current Block Attention and Causal Masking.}  We define the "current block" as the block that contains the query token itself.              
The routing to the current block could also violate causality, since mean pooling across the entire block can inadvertently include information from future tokens.
To address this, we enforce that each token must be routed to its respective current block and apply a causal mask during the current block attention.
This strategy not only avoids any leakage of information from subsequent tokens but also encourages attention to the local context.
More formally, we set $g_i = 1$ for the block $i$ where the position of the query token 
$\mathrm{pos}(\vq)$ is within the interval $I_i$.
From the perspective of Mixture-of-Experts (MoE), the current block attention in MoBA is akin to the role of shared experts in modern MoE architectures~\citep{dai2024deepseekmoe, yang2024qwen2}, where static routing rules are added when expert selection.

Next, we discuss some additional key design choices of MoBA, such as its block segmentation strategy and the hybrid of MoBA and full attention.

\textbf{Fine-Grained Block Segmentation.}
The positive impact of fine-grained expert segmentation in improving mode performance has been well-documented in the Mixture-of-Experts~(MoE) literature~\citep{dai2024deepseekmoe, yang2024qwen2}. In this work, we explore the potential advantage of applying a similar fine-grained segmentation technique to MoBA. MoBA, inspired by MoE,  operates segmentation along the context-length dimension rather than the FFN intermediate hidden dimension. Therefore our investigation aims to determine if MoBA can also benefit when we partition the context into blocks with a finer grain. More experimental results can be found in \Secref{subsec:scaling_ablation}.

\textbf{Hybrid of MoBA and Full Attention.}
MoBA is designed to be a substitute for full attention, maintaining the same number of parameters without any addition or subtraction. This feature inspires us to conduct smooth transitions between full attention and MoBA. Specifically, at the initialization stage, each attention layer has the option to select full attention or MoBA, and this choice can be dynamically altered during training if necessary.
A similar idea of transitioning full attention to sliding window attention has been studied in previous work~\citep{zhang2024simlayerkv}.
More experimental results can be found in \Secref{subsec:hybrid}.

\textbf{Comparing to Sliding Window Attention and Attention Sink.}
Sliding window attention (SWA) and attention sink are two popular sparse attention architectures. We demonstrate that both can be viewed as special cases of MoBA.
For sliding window attention~\citep{beltagy2020longformer}, each query token only attends to its neighboring tokens. This can be interpreted as a variant of MoBA with a gating network that keeps selecting the most recent blocks.
Similarly, attention sink~\citep{xiao2023efficient}, where each query token attends to a combination of initial tokens and the most recent tokens, can be seen as a variant of MoBA with a gating network that always selects both the initial and the recent blocks.
 The above discussion shows that MoBA has stronger expressive power than sliding window attention and attention sink. Moreover, it shows that MoBA can flexibly approximate many static sparse attention architectures by incorporating specific gating networks.

Overall, MoBA's attention mechanism allows the model to adaptively and dynamically focus on the most informative blocks of the context. This is particularly beneficial for tasks involving long documents or sequences, where attending to the entire context may be unnecessary and computationally expensive. MoBA's ability to selectively attend to relevant blocks enables more nuanced and efficient processing of information.

\subsection{Implementation}
\label{subsec:impl}

\begin{algorithm}[t]
\caption{MoBA (Mixture of Block Attention) Implementation}
\label{alg:moba}
\begin{algorithmic}[1]
\REQUIRE{Query, key and value matrices $\mathbf{Q}, \mathbf{K}, \mathbf{V} \in \mathbb{R}^{N\times h\times d}$; MoBA hyperparameters (block size $B$ and top-$k$); $h$ and $d$ denote the number of attention heads and head dimension. Also denote $n=N/B$ to be the number of blocks.}

\STATE // Split KV into blocks
\STATE $\{\tilde{\mathbf{K}}_i, \tilde{\mathbf{V}}_i\} = \text{split\_blocks}(\mathbf{K}, \mathbf{V}, B)$, where $\tilde{\mathbf{K}}_i, \tilde{\mathbf{V}}_i \in \mathbb{R}^{B\times h\times d}, i \in [n]$

\STATE  // Compute gating scores for dynamic block selection
\STATE $\bar{\mathbf{K}} = \text{mean\_pool}(\mathbf{K}, B) \in \mathbb{R}^{n\times h\times d}$
\STATE $\mathbf{S} = \mathbf{Q}\bar{\mathbf{K}}^\top \in \mathbb{R}^{N\times h\times n}$

\STATE // Select blocks with causal constraint (no attention to future blocks)
\STATE $\mathbf{M} = \text{create\_causal\_mask}(N, n)$
\STATE $\mathbf{G} = \text{topk}(\mathbf{S} + \mathbf{M}, k)$

\STATE // Organize attention patterns for computation efficiency
\STATE $\mathbf{Q}^s, \tilde{\mathbf{K}}^s, \tilde{\mathbf{V}}^s = \text{get\_self\_attn\_block}(\mathbf{Q}, \tilde{\mathbf{K}}, \tilde{\mathbf{V}})$
\STATE $\mathbf{Q}^m, \tilde{\mathbf{K}}^m, \tilde{\mathbf{V}}^m = \text{index\_select\_moba\_attn\_block}(\mathbf{Q}, \tilde{\mathbf{K}}, \tilde{\mathbf{V}}, \mathbf{G})$
\STATE // Compute attentions seperately
\STATE $\mathbf{O}^s = \text{flash\_attention\_varlen}(\mathbf{Q}^s, \tilde{\mathbf{K}}^s, \tilde{\mathbf{V}}^s, \text{causal=True})$
\STATE $\mathbf{O}^m = \text{flash\_attention\_varlen}(\mathbf{Q}^m, \tilde{\mathbf{K}}^m, \tilde{\mathbf{V}}^m, \text{causal=False})$

\STATE // Combine results with online softmax
\STATE $\mathbf{O} = \text{combine\_with\_online\_softmax}(\mathbf{O}^s, \mathbf{O}^m)$

\RETURN $\mathbf{O}$
\end{algorithmic}
\end{algorithm}

\begin{figure}[t]
    \centering
    \subfloat[]{
        \includegraphics[width=0.4\textwidth]{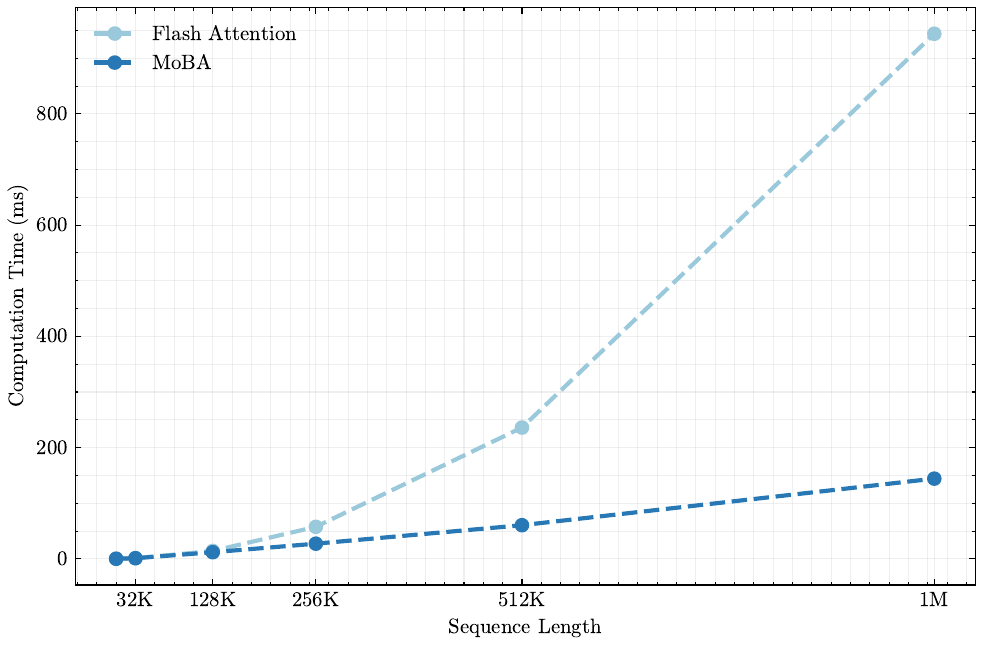}
        \label{fig:1mcompute}
    }
    \subfloat[]{
        \includegraphics[width=0.4\textwidth]{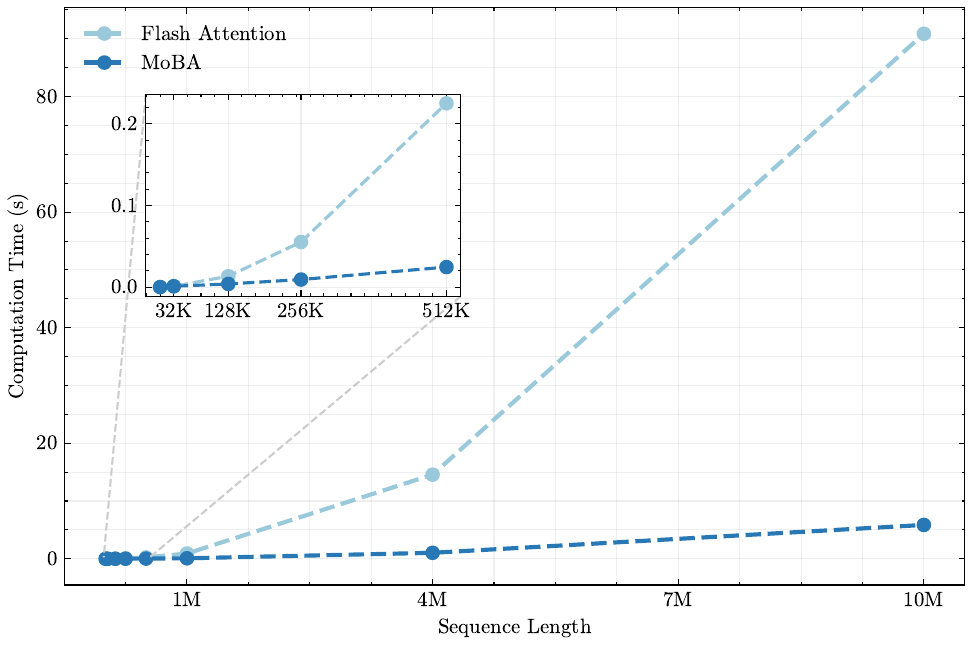}
        \label{fig:10mspeedup}
    }
    \caption{Efficiency of MoBA vs. full attention (implemented with Flash Attention). \textbf{(a)} 1M Model speedup evaluation: Computation time scaling of MoBA versus Flash Attention on 1M model with increasing sequence lengths (8K-1M). \textbf{(b)} Fixed Sparsity Ratio scaling: Computation time scaling comparison between MoBA and Flash Attention across increasing sequence lengths (8K-10M), maintaining a constant sparsity ratio of $95.31\%$ (fixed 64 MoBA blocks with variance block size and fixed top-k=3).
    } 
    \label{Fig:efficiency} 
\end{figure}

We provide a high-performance implementation of MoBA, by incorporating optimization techniques from FlashAttention~\citep{dao2022flashattention} and MoE~\citep{rajbhandari2022deepspeed}. 
\Figref{Fig:efficiency} demonstrates the high efficiency of MoBA, while we defer the detailed experiments on efficiency and scalability to \Secref{subsec:efficiency}.
Our implementation consists of five major steps:

\begin{itemize}
\item Determine the assignment of query tokens to KV blocks according to the gating network and causal mask.
\item Arrange the ordering of query tokens based on their assigned KV blocks.
\item Compute attention outputs for each KV block and the query tokens assigned to it. This step can be optimized by FlashAttention with varying lengths.
\item Re-arrange the attention outputs back to their original ordering.
\item Combine the corresponding attention outputs using online Softmax (i.e., tiling), as a query token may attend to its current block and multiple historical KV blocks.
\end{itemize}

The algorithmic workflow is formalized in Algorithm~\ref{alg:moba} and visualized in \Figref{fig:moba_flash}, illustrating how MoBA can be implemented based on MoE and FlashAttention.
First, the KV matrices are partitioned into blocks (Line 1-2).
Next, the gating score is computed according to \Eqref{eq:gating_score}, which measures the relevance between query tokens and KV blocks (Lines 3-7). 
A top-$k$ operator is applied on the gating score (together with causal mask), resulting in a sparse query-to-KV-block mapping matrix $\mG$ to represent the assignment of queries to KV blocks (Line 8).
Then, query tokens are arranged based on the query-to-KV-block mapping, and block-wise attention outputs are computed (Line 9-12).
Notably, attention to historical blocks (Line 11 and 14) and the current block attention (Line 10 and 13) are computed separately, as additional causality needs to be maintained in the current block attention.
Finally, the attention outputs are rearranged back to their original ordering and combined with online softmax (Line 16)~\citep{milakov2018onlinenormalizercalculationsoftmax,liu2023blockwiseparalleltransformerlarge}.

\section{Experiments}

\subsection{Scaling Law Experiments and Ablation Studies}
\label{subsec:scaling_ablation}

In this section, we conduct scaling law experiments and ablation studies to validate some key design choices of MoBA.

\begin{table}[t]
\centering
\small
\begin{tabular}{c|c|c|c|c|c|c}
\toprule
Model Param & Head & Layer & Hidden & Training Token  & Block size & TopK\\
\midrule
568M & 14 & 14 & 1792 & 10.8B  & 512 & 3 \\
822M & 16 & 16 & 2048 & 15.3B  & 512 & 3 \\
1.1B & 18 & 18 & 2304 & 20.6B  & 512 & 3 \\
1.5B & 20 & 20 & 2560 & 27.4B  & 512 & 3 \\
2.1B & 22 & 22 & 2816 & 36.9B  & 512 & 3 \\
\bottomrule
\end{tabular}
\caption{Configuration of Scaling Law Experiments}
\label{tab:model-specs}
\end{table}

\begin{figure}[t]
    \centering
    \subfloat[]{
        \includegraphics[width=0.45\textwidth]{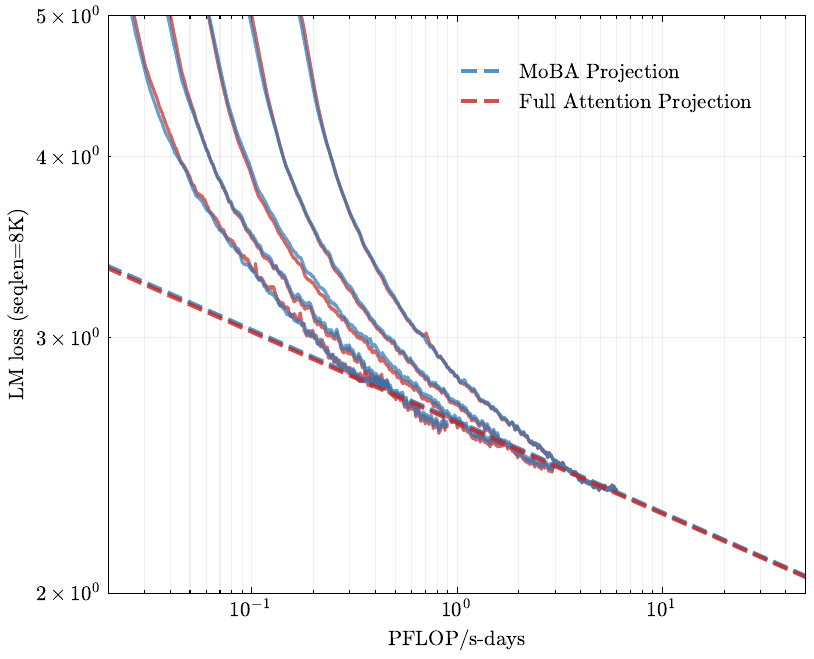}
        \label{fig:scaling_lm_loss}
    }
    \subfloat[]{
        \includegraphics[width=0.45\textwidth]{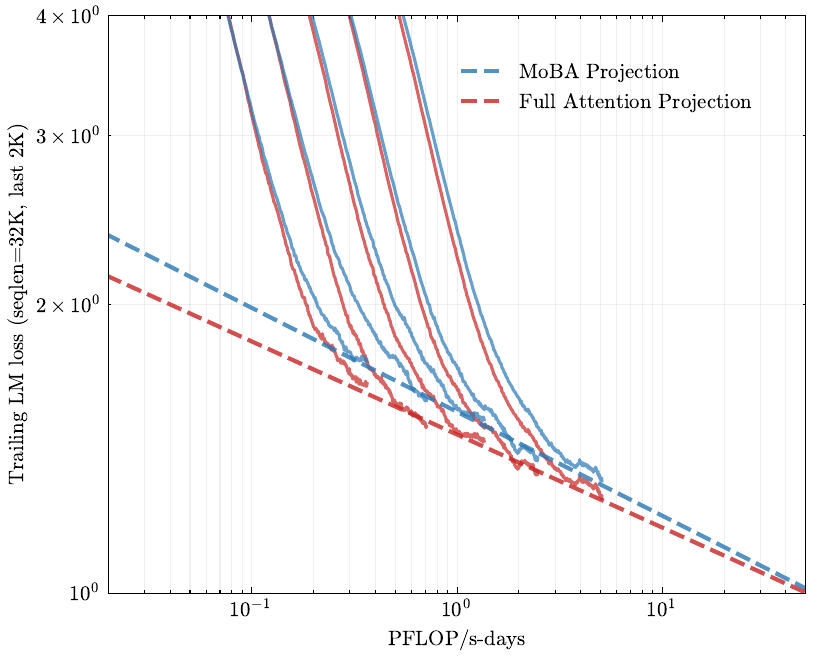}
        \label{fig:scaling_last_block_lm_loss}
    }\\
    \subfloat[]{
\small
\begin{tabular}{c|l|l}
\toprule
L(C) & MoBA & Full \\
\midrule
LM loss (seqlen=8K) & $2.625 \times C^{-0.063}$ & $2.622 \times C^{-0.063}$ \\
\midrule
Trailing LM loss (seqlen=32K, last 2K) & $1.546 \times C^{-0.108}$ & $1.464 \times C^{-0.097}$ \\
\bottomrule
\end{tabular}
\normalsize
\label{tab:comparison}
    }
    \caption{Scaling law comparison between MoBA and full attention. \textbf{(a)} LM loss on validation set (seqlen=8K); \textbf{(b)} trailing LM loss on validation set (seqlen=32K, last 1K tokens); \textbf{(c)} fitted scaling law curve.} 
    \label{Fig:scaling} 
\end{figure}

\paragraph{Scalability w.r.t. LM Loss.} To assess the effectiveness of MoBA, we perform scaling law experiments by comparing the validation loss of language models trained using either full attention or MoBA. Following the Chinchilla scaling law~\citep{hoffmann2022training}, we train five language models of varying sizes with a sufficient number of training tokens to ensure that each model achieves its training optimum.
Detailed configurations of the scaling law experiments can be found in Table~\ref{tab:model-specs}.
Both MoBA and full attention models are trained with a sequence length of 8K. For MoBA models, we set the block size to 512 and select the top-3 blocks for attention, resulting in a sparse attention pattern with sparsity up to $1 - \frac{512\times 3}{8192} = 81.25\%$\footnote{Since we set top-k=3, thus each query token can attend to at most 2 history block and the current block.}.
In particular, MoBA serves as an alternative to full attention, meaning that it does not introduce new parameters or remove existing ones. This design simplifies our comparison process, as the only difference across all experiments lies in the attention modules, while all other hyperparameters, including the learning rate and batch size, remain constant. 
As shown in \Figref{fig:scaling_lm_loss}, the validation loss curves for MoBA and full attention display very similar scaling trends. Specifically, the validation loss differences between these two attention mechanisms remain consistent within a range of $1e-3$. This suggests that MoBA achieves scaling performance that is comparable to full attention, despite its sparse attention pattern with sparsity up to 75\%.

\paragraph{Long Context Scalability.} 
However, LM loss may be skewed by the data length distribution~\citep{an2024does}, which is typically dominated by short sequences. To fully assess the long-context capability of MoBA, we assess the \textbf{LM loss of trailing tokens (trailing LM loss, in short)}, which computes the LM loss of the last few tokens in the sequence. We count this loss only for sequences that reach the maximum sequence length to avoid biases that may arise from very short sequences. A detailed discussion on trailing tokens scaling can be found in the Appendix~\ref{subsec:trailing_scaling}

These metrics provide insights into the model's ability to generate the final portion of a sequence, which can be particularly informative for tasks involving long context understanding.
Therefore, we adopt a modified experimental setting by increasing the maximum sequence length from 8k to 32k. 
This adjustment leads to an even sparser attention pattern for MoBA, achieving a sparsity level of up to 
$1 - \frac{512\times 3}{32768} = 95.31\%$.  
As shown in \Figref{fig:scaling_last_block_lm_loss}, although MoBA exhibits a marginally higher last block LM loss compared to full attention in all five experiments, the loss gap is progressively narrowing. This experiment implies the long-context scalability of MoBA.

\begin{figure}[htbp] 
    \centering 
    \includegraphics[width=0.4\textwidth]{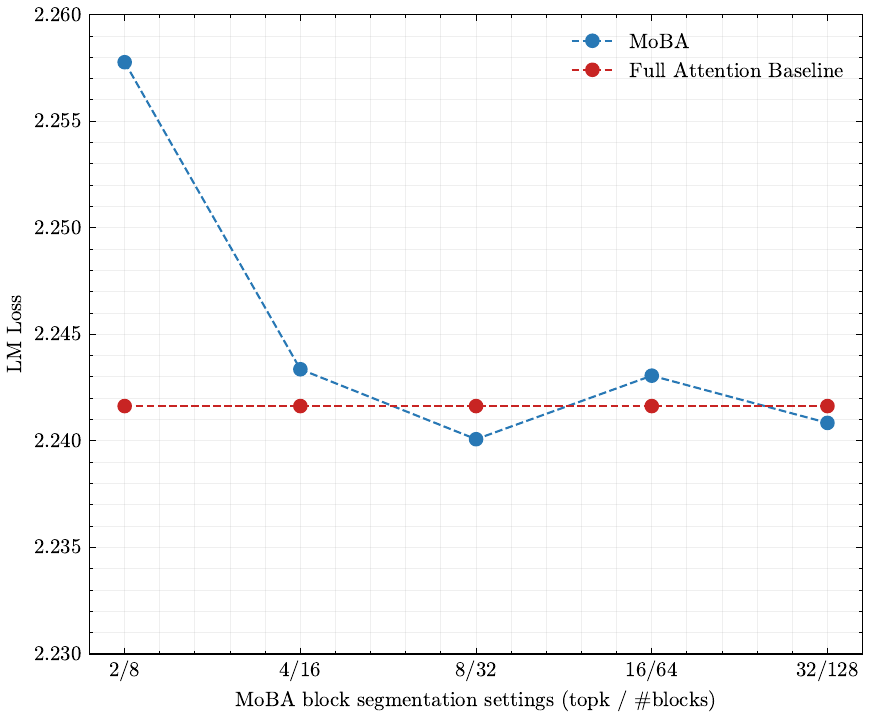} 
    \caption{Fine-Grained Block Segmentation. The LM loss on validation set v.s. MoBA with different block granularity.} 
    \label{fig:fine-grain} 
\end{figure}

\paragraph{Ablation Study on Fine-Grained Block Segmentation.} 
We further ablate the block granularity of MoBA. We carry out a series of experiments using a 1.5B parameter model with a 32K context length. The hyperparameters of block size and top-k are adjusted to maintain a consistent level of attention sparsity. Specifically, we divide the 32K context into 8, 16, 32, 64, and 128 blocks, and correspondingly select 2, 4, 8, 16, and 32 blocks, ensuring an attention sparsity of 75\% across these configurations. 
As shown in \Figref{fig:fine-grain}, MoBA's performance is significantly affected by block granularity. Specifically, there is a performance difference of 1e-2 between the coarsest-grained setting (selecting 2 blocks from 8) and the settings with finer granularity.
These findings suggest that fine-grained segmentation appears to be a general technique for enhancing the performance of models within the MoE family, including MoBA.

\subsection{Hybrid of MoBA and Full Attention}
\label{subsec:hybrid}

\begin{figure}[t]
    \centering
    \subfloat[]{
        \includegraphics[width=0.27\textwidth]{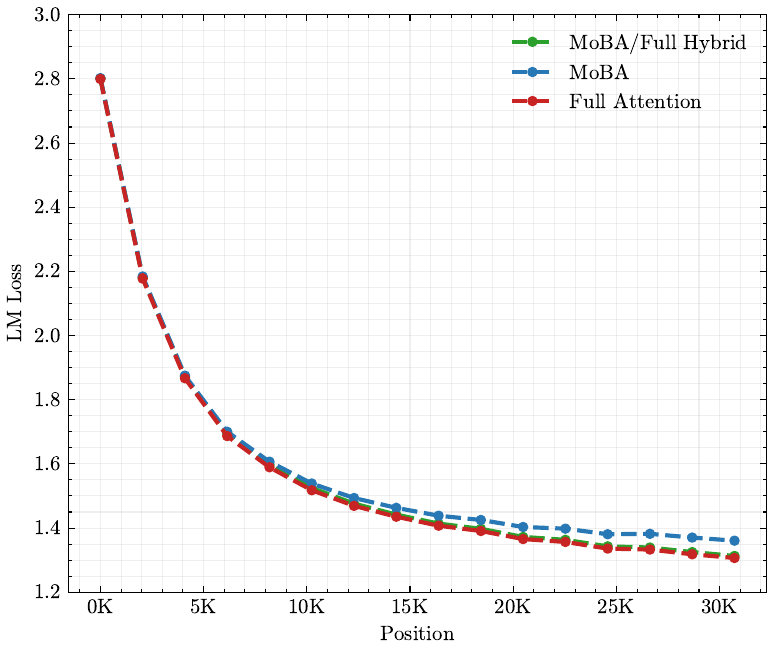} 
        \label{fig:hybrid_training} 
    }
    \subfloat[]{
        \includegraphics[width=0.3\textwidth]{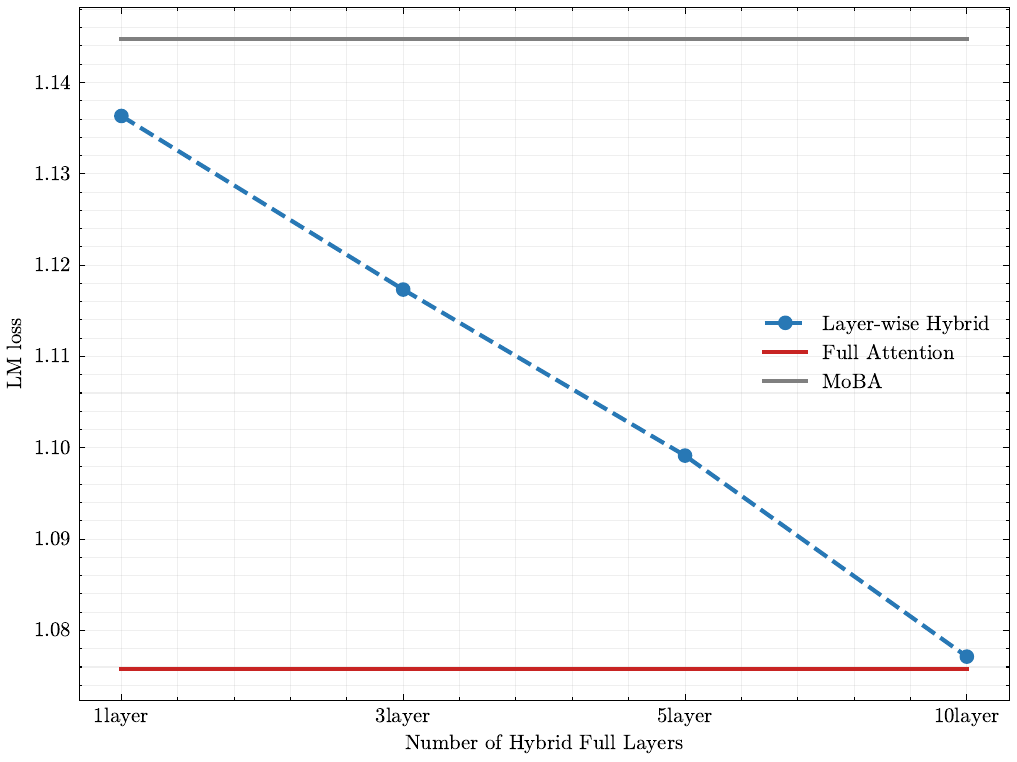}
        \label{fig:sft_loss_overall}
    }
    \subfloat[]{
        \includegraphics[width=0.3\textwidth]{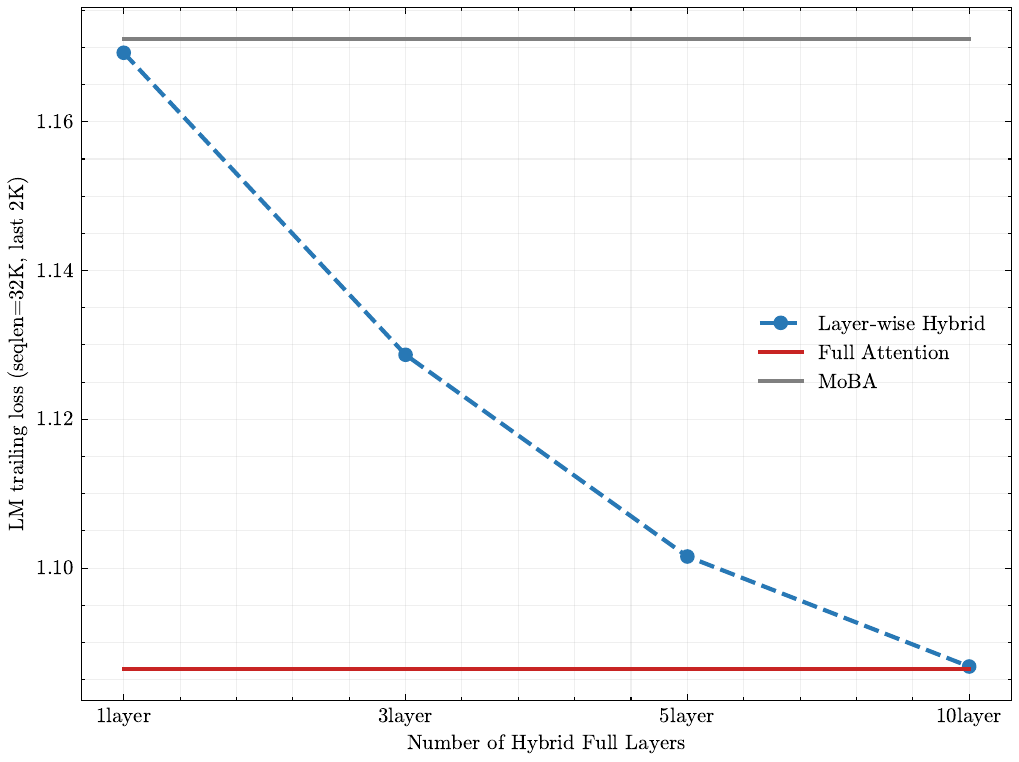}
        \label{fig:sft_loss_last_block}
    }
    \caption{Hybrid of MoBA and full attention. \textbf{(a)}
    position-wise LM loss for MoBA, full attention, and MoBA/full hybrid training;
    \textbf{(b)} SFT LM loss w.r.t the number of full attention layers in layer-wise hybrid;
    \textbf{(c)} SFT trailing LM loss (seqlen=32K, last 2K) w.r.t the number of full attention layers in layer-wise hybrid.
    } 
\end{figure}

As discussed in \Secref{sec:method}, we design MoBA to be a flexible substitute for full attention, so that it can easily switch from/to full attention with minimal overhead and achieve comparable long-context performance. 
In this section, we first show seamless transition between full attention and MoBA can be a solution for efficient long-context pre-training. Then we discuss the layer-wise hybrid strategy, mainly for the performance of supervised fine-tuning~(SFT).

\paragraph{MoBA/Full Hybrid Training.}
 We train three models, each with 1.5B parameters, on 30B tokens with a context length of 32K tokens.
 For the hyperparameters of MoBA, the block size is set to 2048, and the top-k parameter is set to 3.
  The detailed training recipes are as follows:
\begin{itemize}
\item MoBA/full hybrid: This model is trained using a two-stage recipe. In the first stage, MoBA is used to train on 90\% of the tokens. In the second stage, the model switches to full attention for the remaining 10\% of the tokens.
\item Full attention: This model is trained using full attention throughout the entire training.
\item MoBA: This model is trained exclusively using MoBA.
\end{itemize}
We evaluate their long-context performance via position-wise language model~(LM) loss, which is a fine-grained metric to evaluate lm loss at each position within a sequence. Unlike the vanilla LM loss, which is computed by averaging the LM loss across all positions, the position-wise LM loss breaks down the loss for each position separately. 
Similar metrics have been suggested by previous studies~\citep{xiong2023effectivelongcontextscalingfoundation,reid2024gemini}, who noticed that position-wise LM loss follows a power-law trend relative to context length. 
As shown in \Figref{fig:hybrid_training}, the MoBA-only recipe results in higher position-wise losses for trailing tokens.  
Importantly, our MoBA/full hybrid recipe reaches a loss nearly identical to that of full attention. This result highlights the effectiveness of the MoBA/full hybrid training recipe in balancing training efficiency with model performance. More interestingly, we have not observed significant loss spikes during the switch between MoBA and full attention, again demonstrating the flexibility and robustness of MoBA. 

\paragraph{Layer-wise Hybrid.} This flexibility of MoBA  encourages us to delve into a more sophisticated strategy --- the layer-wise hybrid of MoBA and full attention. We investigate this strategy with a particular focus on its application during the supervised fine-tuning (SFT).
The motivation for investigating this strategy stems from our observation that MoBA sometimes results in suboptimal performance during SFT, as shown in \Figref{fig:sft_loss_overall}. We speculate that this may be attributed to the loss masking employed in SFT --- prompt tokens are typically excluded from the loss calculation during SFT, which can pose a sparse gradient challenge for sparse attention methods like MoBA.
Because it may hinder the backpropagation of gradients, which are initially calculated from unmasked tokens, throughout the entire context.
To address this issue, we propose a hybrid approach --- switching the last several Transformer layers from MoBA to full attention, while the remaining layers continue to employ MoBA. 
 As shown in \Figref{fig:sft_loss_overall} and \Figref{fig:sft_loss_last_block}, this strategy can significantly reduce SFT loss.

\subsection{Large Language Modeling Evaluation}
\label{subsec:llama8B}

\begin{figure}[t] 
    \centering 
    \includegraphics[width=0.8\textwidth]{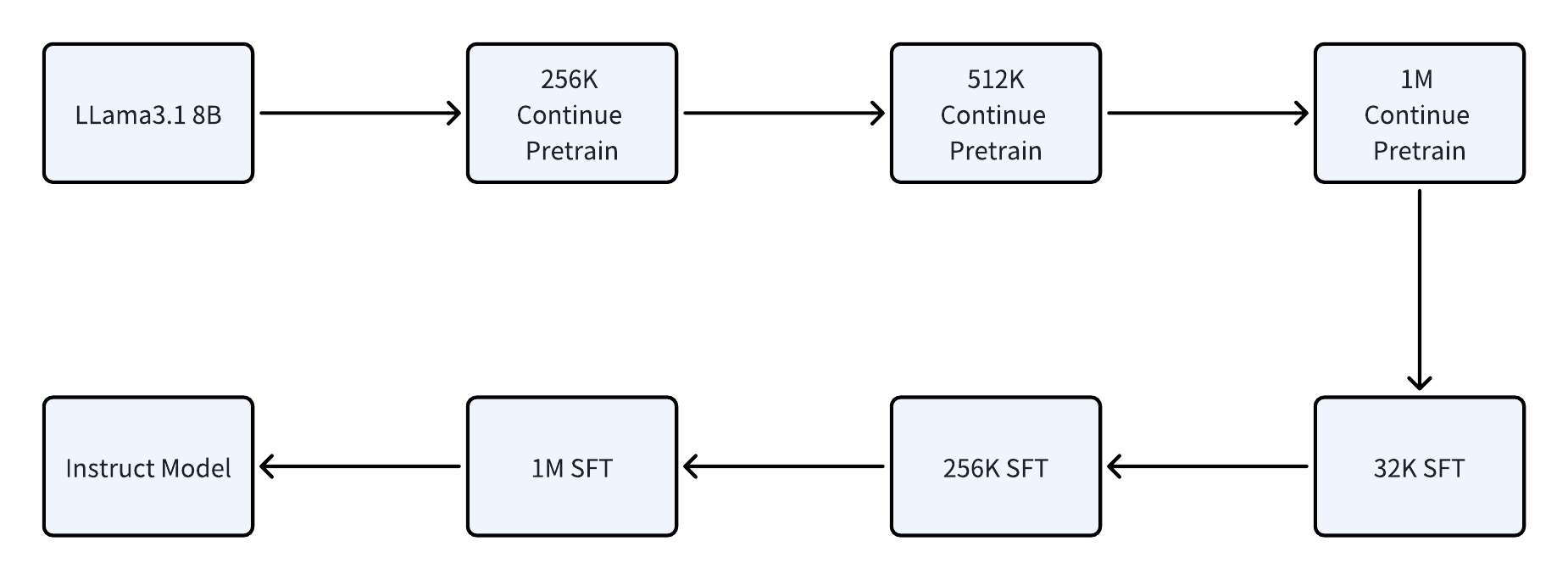} 
    \caption{The continual pre-training and SFT recipes.}
    \label{fig:recipe} 
\end{figure}

 We conduct a thorough assessment of MoBA across a variety of real-world downstream tasks, evaluating its performance in comparison to full attention models. 
For ease of verification, our experiments begin with the Llama 3.1 8B Base Model, which is used as the starting point for long-context pre-training.
 This model, termed Llama-8B-1M-MoBA, is initially trained with a context length of 128K tokens, and we gradually increase the context length to 256K, 512K, and 1M tokens during the continual pre-training. To ease this transition, we use  position interpolation method~\citep{chen2023extendingcontextwindowlarge} at the start of the 256K continual pre-training stage. This technique enables us to extend the effective context length from 128K tokens to 1M tokens.
After completing the 1M continuous pre-training, MoBA is activated for 100B tokens. We set the block size to 4096 and the top-K parameter to 12, leading to an attention sparsity of up to $1-\frac{4096\times 12}{1M}=95.31\%$. 
To preserve some full attention capabilities, we adopt the layer-wise hybrid strategy --- the last three layers remain as full attention, while the other 29 full attention layers are switched to MoBA.
For supervised fine-tuning, we follow a similar strategy that gradually increases the context length from 32K to 1M. 
The baseline full attention models (termed Llama-8B-1M-Full) also follow a similar training strategy as shown in \Figref{fig:recipe}, with the only difference being the use of full attention throughout the process. This approach allows us to directly compare the performance of MoBA with that of full attention models under equivalent training conditions.

\begin{figure}[t] 
    \centering 
    \includegraphics[width=0.9\textwidth]{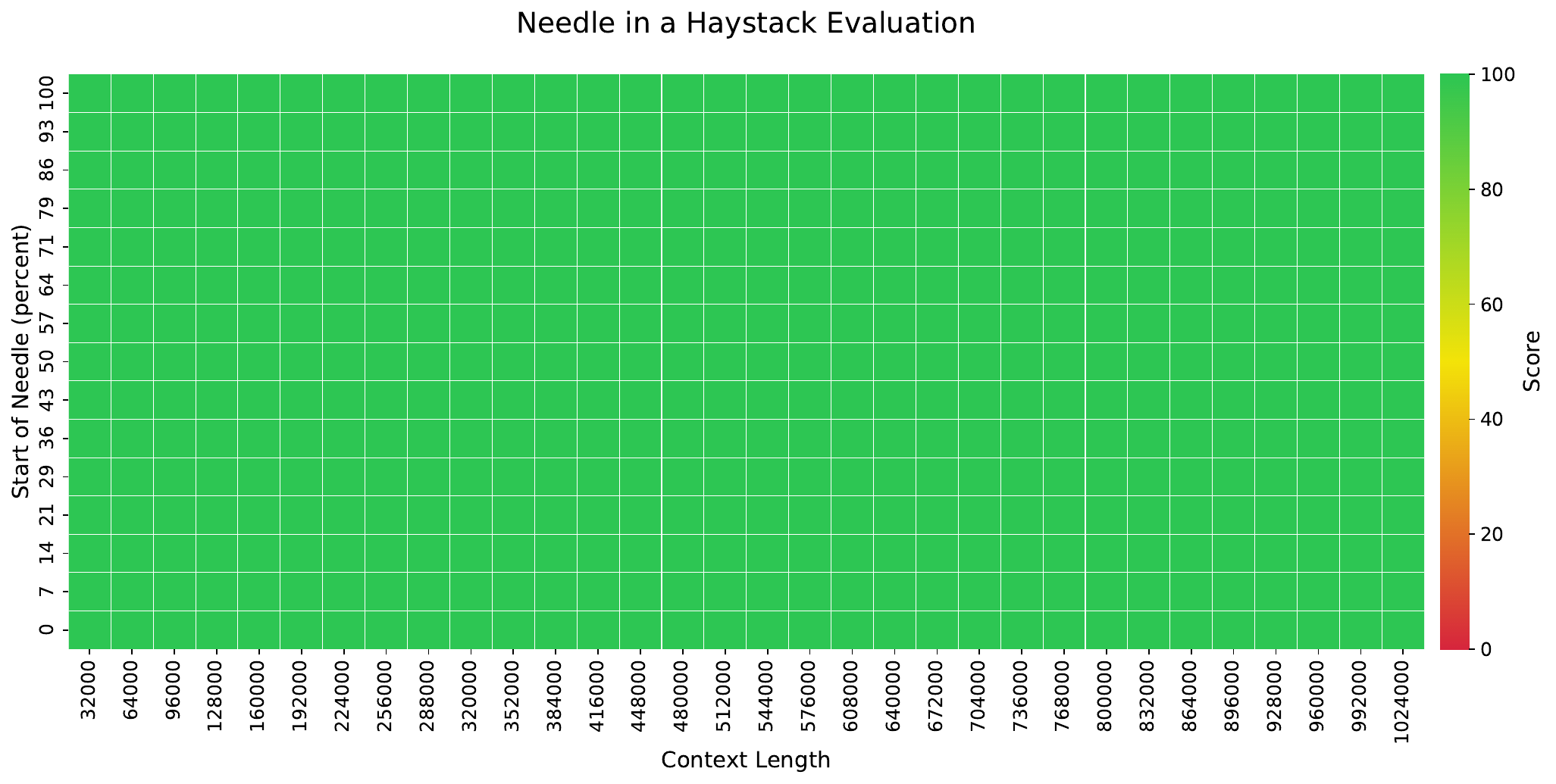} 
    \caption{Performance of LLama-8B-1M-MoBA on the Needle in the Haystack benchmark~(upto 1M context length).} 
    \label{fig:needle} 
\end{figure}

The evaluation is performed on several widely used long-context benchmarks. 
In particular, across all evaluation tasks, MoBA is used for prefill only, while we switch to full attention during generation for better performance.
As shown in Table~\ref{tab:bench}, Llama-8B-1M-MoBA exhibits a performance that is highly comparable to that of Llama-8B-1M-Full.
It is particularly noteworthy that in the longest benchmark, RULER, where MoBA operates at a sparsity level of up to $1-\frac{4096\times 12}{128K}=62.5\%$, Llama-8B-1M-MoBA nearly matches the performance of Llama-8B-1M-Full, with a score of 0.7818 compared to 0.7849.
For context lengths of up to 1M tokens, we evaluate the model using the traditional Needle in the Haystack benchmark. As shown in \Figref{fig:needle}, Llama-8B-1M-MoBA demonstrates satisfactory performance even with an extended context length of 1 million tokens.

\begin{table}[t]
\centering
\small
\begin{tabular}{l|c|c}
\toprule
\textbf{Benchmark} & \textbf{Llama-8B-1M-MoBA} & \textbf{Llama-8B-1M-Full} \\
\midrule
AGIEval [0-shot] & 0.5144 & \textbf{0.5146} \\
BBH [3-shot] & 0.6573 & \textbf{0.6589} \\
CEval [5-shot] & \textbf{0.6273} & 0.6165 \\
GSM8K [5-shot] & \textbf{0.7278} & 0.7142 \\
HellaSWAG [0-shot] & 0.8262 & \textbf{0.8279} \\
Loogle [0-shot] & \textbf{0.4209} & 0.4016 \\
Competition Math [0-shot] & 0.4254 & \textbf{0.4324} \\
MBPP [3-shot] & \textbf{0.5380} & 0.5320 \\
MBPP Sanitized [0-shot] & \textbf{0.6926} & 0.6615 \\
MMLU [0-shot] & 0.4903 & \textbf{0.4904} \\
MMLU Pro [5-shot][CoT] & 0.4295 & \textbf{0.4328} \\
OpenAI HumanEval [0-shot][pass@1] & 0.6951 & \textbf{0.7012} \\
SimpleQA [0-shot] & 0.0465 & \textbf{0.0492} \\
TriviaQA [0-shot] & \textbf{0.5673} & 0.5667 \\
LongBench @32K [0-shot] & \textbf{0.4828} & 0.4821 \\
RULER @128K [0-shot] & 0.7818 & \textbf{0.7849} \\
\bottomrule
\end{tabular}
\caption{Performance comparison between MoBA and full Attention across different evaluation benchmarks.}
\label{tab:bench}
\end{table}

\subsection{Efficiency and Scalability}
\label{subsec:efficiency}

The above experimental results show that MoBA achieves comparable performance not only regarding language model losses but also in real-world tasks.
To further investigate its efficiency, we compare the forward pass time of the attention layer in two models trained in \Secref{subsec:llama8B} ---  Llama-8B-1M-MoBA and Llama-8B-1M-Full. We focus solely on the attention layer, as all other layers (e.g., FFN) have identical FLOPs in both models.
As shown in \Figref{fig:1mcompute}, MoBA is more efficient than full attention across all context lengths, demonstrating a  sub-quadratic computational complexity. In particular, it achieves a speedup ratio of up to 6.5x when prefilling 1M tokens.

We also explore the length scalability of MoBA by gradually increasing the context length to 10 million tokens. To maintain a constant attention sparsity, we keep the top-k value and number of MoBA Block fixed while proportionally increasing the block size.
To reach the 10M context length, we expanded tensor parallelism \citep{shoeybi2019megatron} toward the query head level, Specifically, we broadcast key and value tensors across distributed query heads, effectively addressing GPU memory limitations while preserving computational efficiency.
As shown in \Figref{fig:10mspeedup}, MoBA demonstrates superior efficiency compared to standard Flash Attention when scaling to longer sequences. Specifically, at 10M tokens moba achieves a speedup ratio of 16x reduction in attention computation time.
The inset graph in the top figure, focusing on shorter sequences (32K to 512K), shows that even though both methods perform comparably at smaller scales, MoBA's computational advantage becomes increasingly evident as sequences grow longer, highlighting its particular strength in processing extremely long sequences.

Overall, the high efficiency of MoBA can be attributed to two key innovations: (1) the block sparse attention mechanism, and (2) the optimized implementation combining Mixture-of-Experts (MoE) and FlashAttention, as described in \Secref{subsec:impl}. These techniques effectively address the quadratic complexity limitation of full attention, reducing the computational complexity to a more economical sub-quadratic scale.

\section{Related Work}

The development of efficient attention~\citep{tay2020efficient} mechanisms has been a critical area of research in the field of natural language processing, particularly with the rise of Large Language Models (LLMs). As the demand for handling longer sequences and reducing computational costs grows, efficeint attention techniques have emerged as a promising solution to reduce the quadratic complexity of self-attention mechanisms while maintaining model performance.

\textbf{Static Sparse Patterns:}
Significant efforts, such as Sparse Transformer~\citep{child2019generating}, Star-Transformer~\citep{guo2019star}, BlockBERT~\citep{qiu2019blockwise}, Longformer~\citep{beltagy2020longformer},
GMAT~\citep{gupta2020gmat}, ETC~\citep{ainslie2020etc},
BigBird~\citep{zaheer2020big}, LongT5~\citep{guo2021longt5} and LongNet~\citep{ding2023longnet}, have been dedicated to the design of static attention patterns in LLMs. 
Their choices of static attention patterns can encompass strided and fixed attention, window attention, global token attention, random attention, dilated attention, block sparse attention, or any combinations of them. 
In the realm of multimodal models, static sparse attention mechanisms have also been developed, such as axial attention~\citep{ho2019axial} for 2D images and spatial-temporal attention~\citep{opensora} for 3D videos.

\textbf{Dynamic Sparse Patterns:}
Different from static patterns, dynamic sparse attention techniques adaptively determine which tokens to attend. 
Reformer~\citep{kitaev2020reformer} and Routing Transformer~\citep{roy2021efficient} respectively employ locality-sensitive hashing~(LSH) and K-means to cluster tokens, and attend to clusters rather than the full context. Memorizing Transformers~\citep{wu2022memorizing} and Unlimiformer~\citep{bertsch2024unlimiformer} dynamically attend to tokens selected by the k-nearest-neighbor~(kNN) algorithms. 
CoLT5~\citep{ainslie2023colt5} designs a routing modules to select the most important queries and keys.
Sparse Sinkhorn Attention~\citep{tay2020sparse} learns to permute blocks from the input sequence, allowing dynamic block sparse attention computation.

\textbf{Training-free Sparse Attention:}
In addition to the previously discussed approaches that study training sparse attention models, there are also strategies designed to incorporate sparse attention mechanisms to enhance the efficiency of the two primary stages of model inference --- either the prefill stage or the decode stage, or both of them.
During the prefill optimization phase, the complete prompt can be utilized for attention profiling, which allows for the exploration of more intricate sparse attention patterns. For instance, MoA~\citep{fu2024moa}, Minference~\citep{jiang2024minference}, and SeerAttention~\citep{gao2024seerattention} have investigated sparse attention configurations such as A-shape, vertical-slash, and dynamic block sparsity.
In the context of decode optimization, considerable work has been dedicated to compressing and pruning the KV-cache to achieve a balance between the quality and speed of text generation. Notable efforts in this area include H2O~\citep{zhang2024h2o}, StreamingLLM~\citep{xiao2023efficient}, TOVA~\citep{oren2024tova}, FastGen~\citep{ge2023fastgen} and  Quest~\citep{tang2024quest}. Quest, in particular, can be viewed as MoBA with a smaller block size and a specialized block representation function which combines both min and max pooling. Another work closely related to MoBA is Longheads~\citep{lu2024longheads} which can be viewed as MoBA with a top-1 gating network, meaning that each query selects the most relevant KV blocks for attention.

\textbf{Beyond Traditional Attention Architecture:} Another line of research  investigates novel model architectures that deviate from the conventional attention mechanism. As architectures change, these methods require training models from scratch and are unable to reuse pre-trained Transformer-based models. 
 Studies in this domain have explored architectures that are inspired by  Convolutional Neural Networks (CNNs), Recurrent Neural Networks (RNNs), State Space Models (SSMs), or Linear Attention~\citep{katharopoulos2020transformers}, Examples of such models include Hyena~\citep{poli2023hyena},  Performer~\citep{choromanski2020rethinking},
Linformer~\citep{wang2020linformer},
RWKV~\cite{peng2023rwkv}, Mamba~\citep{gu2023mamba}, RetNet~\citep{sun2023retentive}, etc.

In summary, the landscape of efficient attention techniques is diverse, encompassing sparse patterns that range from static to dynamic, optimization objectives that span from training to inference, and architectures that extend from traditional attention mechanisms to innovative alternatives.
Each method presents unique advantages and trade-offs, and the choice of technique often depends on the specific requirements of the application, such as the maximum sequence length, computational resources, and the desired balance between efficiency and performance. As research in this area continues to evolve, it is expected that these methods will play a crucial role in enabling LLMs to tackle increasingly complex tasks while maintaining efficiency and scalability.

\section{Conclusion}

In this paper, we introduce Mixture of Block Attention (MoBA), a novel attention architecture inspired by the principles of Mixture of Experts (MoE) that aims to enhance the efficiency and scalability of large language models (LLMs) for long-context tasks. MoBA addresses the computational challenges associated with traditional attention mechanisms by partitioning the context into blocks and employing a dynamic gating mechanism to selectively route query tokens to the most relevant KV blocks. This approach not only reduces computational complexity but also maintains model performance. Moreover, it allows for seamless transitions between full and sparse attention.
Through extensive experiments, we demonstrated that MoBA achieves performance comparable to full attention while significantly improving computational efficiency. Our results show that MoBA can scale effectively to long contexts, maintaining low LM losses and high performance on various benchmarks. Additionally, MoBA's flexibility allows it to be integrated with existing models without substantial training cost, making it a practical continual pre-training solution for enhancing long-context capabilities in LLMs.
In summary, MoBA represents a significant advancement in efficient attention, offering a balanced approach between performance and efficiency. Future work may explore further optimizations of MoBA's block-selection strategies, investigate its application to other modalities, and study its potential for improving generalization in complex reasoning tasks.

\printbibliography[title={References}]

@article{child2019generating,
  title={Generating long sequences with sparse transformers},
  author={Child, Rewon and Gray, Scott and Radford, Alec and Sutskever, Ilya},
  journal={arXiv preprint arXiv:1904.10509},
  year={2019}
}

@article{guo2021longt5,
  title={LongT5: Efficient text-to-text transformer for long sequences},
  author={Guo, Mandy and Ainslie, Joshua and Uthus, David and Ontanon, Santiago and Ni, Jianmo and Sung, Yun-Hsuan and Yang, Yinfei},
  journal={arXiv preprint arXiv:2112.07916},
  year={2021}
}

@article{zaheer2020big,
  title={Big bird: Transformers for longer sequences},
  author={Zaheer, Manzil and Guruganesh, Guru and Dubey, Kumar Avinava and Ainslie, Joshua and Alberti, Chris and Ontanon, Santiago and Pham, Philip and Ravula, Anirudh and Wang, Qifan and Yang, Li and others},
  journal={Advances in neural information processing systems},
  volume={33},
  pages={17283--17297},
  year={2020}
}

@article{ding2023longnet,
  title={Longnet: Scaling transformers to 1,000,000,000 tokens},
  author={Ding, Jiayu and Ma, Shuming and Dong, Li and Zhang, Xingxing and Huang, Shaohan and Wang, Wenhui and Zheng, Nanning and Wei, Furu},
  journal={arXiv preprint arXiv:2307.02486},
  year={2023}
}

@article{ho2019axial,
  title={Axial attention in multidimensional transformers},
  author={Ho, Jonathan and Kalchbrenner, Nal and Weissenborn, Dirk and Salimans, Tim},
  journal={arXiv preprint arXiv:1912.12180},
  year={2019}
}

@software{opensora,
  author = {Zangwei Zheng and Xiangyu Peng and Tianji Yang and Chenhui Shen and Shenggui Li and Hongxin Liu and Yukun Zhou and Tianyi Li and Yang You},
  title = {Open-Sora: Democratizing Efficient Video Production for All},
  month = {March},
  year = {2024},
  url = {https://github.com/hpcaitech/Open-Sora}
}

@article{beltagy2020longformer,
  title={Longformer: The long-document transformer},
  author={Beltagy, Iz and Peters, Matthew E and Cohan, Arman},
  journal={arXiv preprint arXiv:2004.05150},
  year={2020}
}

@article{ainslie2020etc,
  title={ETC: Encoding long and structured inputs in transformers},
  author={Ainslie, Joshua and Ontanon, Santiago and Alberti, Chris and Cvicek, Vaclav and Fisher, Zachary and Pham, Philip and Ravula, Anirudh and Sanghai, Sumit and Wang, Qifan and Yang, Li},
  journal={arXiv preprint arXiv:2004.08483},
  year={2020}
}

@article{gupta2020gmat,
  title={Gmat: Global memory augmentation for transformers},
  author={Gupta, Ankit and Berant, Jonathan},
  journal={arXiv preprint arXiv:2006.03274},
  year={2020}
}

@article{guo2019star,
  title={Star-transformer},
  author={Guo, Qipeng and Qiu, Xipeng and Liu, Pengfei and Shao, Yunfan and Xue, Xiangyang and Zhang, Zheng},
  journal={arXiv preprint arXiv:1902.09113},
  year={2019}
}

@article{qiu2019blockwise,
  title={Blockwise self-attention for long document understanding},
  author={Qiu, Jiezhong and Ma, Hao and Levy, Omer and Yih, Scott Wen-tau and Wang, Sinong and Tang, Jie},
  journal={arXiv preprint arXiv:1911.02972},
  year={2019}
}

@article{kitaev2020reformer,
  title={Reformer: The efficient transformer},
  author={Kitaev, Nikita and Kaiser, {\L}ukasz and Levskaya, Anselm},
  journal={arXiv preprint arXiv:2001.04451},
  year={2020}
}

@article{roy2021efficient,
  title={Efficient content-based sparse attention with routing transformers},
  author={Roy, Aurko and Saffar, Mohammad and Vaswani, Ashish and Grangier, David},
  journal={Transactions of the Association for Computational Linguistics},
  volume={9},
  pages={53--68},
  year={2021},
  publisher={MIT Press One Rogers Street, Cambridge, MA 02142-1209, USA journals-info~…}
}

@article{wu2022memorizing,
  title={Memorizing transformers},
  author={Wu, Yuhuai and Rabe, Markus N and Hutchins, DeLesley and Szegedy, Christian},
  journal={arXiv preprint arXiv:2203.08913},
  year={2022}
}

@article{bertsch2024unlimiformer,
  title={Unlimiformer: Long-range transformers with unlimited length input},
  author={Bertsch, Amanda and Alon, Uri and Neubig, Graham and Gormley, Matthew},
  journal={Advances in Neural Information Processing Systems},
  volume={36},
  year={2024}
}

@article{ainslie2023colt5,
  title={Colt5: Faster long-range transformers with conditional computation},
  author={Ainslie, Joshua and Lei, Tao and de Jong, Michiel and Onta{\~n}{\'o}n, Santiago and Brahma, Siddhartha and Zemlyanskiy, Yury and Uthus, David and Guo, Mandy and Lee-Thorp, James and Tay, Yi and others},
  journal={arXiv preprint arXiv:2303.09752},
  year={2023}
}

@inproceedings{tay2020sparse,
  title={Sparse sinkhorn attention},
  author={Tay, Yi and Bahri, Dara and Yang, Liu and Metzler, Donald and Juan, Da-Cheng},
  booktitle={International Conference on Machine Learning},
  pages={9438--9447},
  year={2020},
  organization={PMLR}
}

@article{zhang2024h2o,
  title={H2o: Heavy-hitter oracle for efficient generative inference of large language models},
  author={Zhang, Zhenyu and Sheng, Ying and Zhou, Tianyi and Chen, Tianlong and Zheng, Lianmin and Cai, Ruisi and Song, Zhao and Tian, Yuandong and R{\'e}, Christopher and Barrett, Clark and others},
  journal={Advances in Neural Information Processing Systems},
  volume={36},
  year={2024}
}

@article{xiao2023efficient,
  title={Efficient streaming language models with attention sinks},
  author={Xiao, Guangxuan and Tian, Yuandong and Chen, Beidi and Han, Song and Lewis, Mike},
  journal={arXiv preprint arXiv:2309.17453},
  year={2023}
}

@article{jiang2024minference,
  title={Minference 1.0: Accelerating pre-filling for long-context llms via dynamic sparse attention},
  author={Jiang, Huiqiang and Li, Yucheng and Zhang, Chengruidong and Wu, Qianhui and Luo, Xufang and Ahn, Surin and Han, Zhenhua and Abdi, Amir H and Li, Dongsheng and Lin, Chin-Yew and others},
  journal={arXiv preprint arXiv:2407.02490},
  year={2024}
}

@article{fu2024moa,
  title={Moa: Mixture of sparse attention for automatic large language model compression},
  author={Fu, Tianyu and Huang, Haofeng and Ning, Xuefei and Zhang, Genghan and Chen, Boju and Wu, Tianqi and Wang, Hongyi and Huang, Zixiao and Li, Shiyao and Yan, Shengen and others},
  journal={arXiv preprint arXiv:2406.14909},
  year={2024}
}

@article{gao2024seerattention,
  title={SeerAttention: Learning Intrinsic Sparse Attention in Your LLMs},
  author={Gao, Yizhao and Zeng, Zhichen and Du, Dayou and Cao, Shijie and So, Hayden Kwok-Hay and Cao, Ting and Yang, Fan and Yang, Mao},
  journal={arXiv preprint arXiv:2410.13276},
  year={2024}
}

@article{ge2023fastgen,
  title={Model tells you what to discard: Adaptive kv cache compression for llms},
  author={Ge, Suyu and Zhang, Yunan and Liu, Liyuan and Zhang, Minjia and Han, Jiawei and Gao, Jianfeng},
  journal={arXiv preprint arXiv:2310.01801},
  year={2023}
}

@article{tang2024quest,
  title={Quest: Query-Aware Sparsity for Efficient Long-Context LLM Inference},
  author={Tang, Jiaming and Zhao, Yilong and Zhu, Kan and Xiao, Guangxuan and Kasikci, Baris and Han, Song},
  journal={arXiv preprint arXiv:2406.10774},
  year={2024}
}

@article{oren2024tova,
  title={Transformers are multi-state rnns},
  author={Oren, Matanel and Hassid, Michael and Adi, Yossi and Schwartz, Roy},
  journal={arXiv preprint arXiv:2401.06104},
  year={2024}
}

@article{gu2023mamba,
  title={Mamba: Linear-time sequence modeling with selective state spaces},
  author={Gu, Albert and Dao, Tri},
  journal={arXiv preprint arXiv:2312.00752},
  year={2023}
}

@article{peng2023rwkv,
  title={Rwkv: Reinventing rnns for the transformer era},
  author={Peng, Bo and Alcaide, Eric and Anthony, Quentin and Albalak, Alon and Arcadinho, Samuel and Biderman, Stella and Cao, Huanqi and Cheng, Xin and Chung, Michael and Grella, Matteo and others},
  journal={arXiv preprint arXiv:2305.13048},
  year={2023}
}

@inproceedings{poli2023hyena,
  title={Hyena hierarchy: Towards larger convolutional language models},
  author={Poli, Michael and Massaroli, Stefano and Nguyen, Eric and Fu, Daniel Y and Dao, Tri and Baccus, Stephen and Bengio, Yoshua and Ermon, Stefano and R{\'e}, Christopher},
  booktitle={International Conference on Machine Learning},
  pages={28043--28078},
  year={2023},
  organization={PMLR}
}

@article{choromanski2020rethinking,
  title={Rethinking attention with performers},
  author={Choromanski, Krzysztof and Likhosherstov, Valerii and Dohan, David and Song, Xingyou and Gane, Andreea and Sarlos, Tamas and Hawkins, Peter and Davis, Jared and Mohiuddin, Afroz and Kaiser, Lukasz and others},
  journal={arXiv preprint arXiv:2009.14794},
  year={2020}
}

@article{wang2020linformer,
  title={Linformer: Self-attention with linear complexity},
  author={Wang, Sinong and Li, Belinda Z and Khabsa, Madian and Fang, Han and Ma, Hao},
  journal={arXiv preprint arXiv:2006.04768},
  year={2020}
}

@inproceedings{katharopoulos2020transformers,
  title={Transformers are rnns: Fast autoregressive transformers with linear attention},
  author={Katharopoulos, Angelos and Vyas, Apoorv and Pappas, Nikolaos and Fleuret, Fran{\c{c}}ois},
  booktitle={International conference on machine learning},
  pages={5156--5165},
  year={2020},
  organization={PMLR}
}

@article{tay2020efficient,
  title={Efficient transformers: A survey. arXiv},
  author={Tay, Yi and Dehghani, Mostafa and Bahri, Dara and Metzler, Donald},
  journal={arXiv preprint arXiv:2009.06732},
  year={2020}
}

@article{reid2024gemini,
  title={Gemini 1.5: Unlocking multimodal understanding across millions of tokens of context},
  author={Reid, Machel and Savinov, Nikolay and Teplyashin, Denis and Lepikhin, Dmitry and Lillicrap, Timothy and Alayrac, Jean-baptiste and Soricut, Radu and Lazaridou, Angeliki and Firat, Orhan and Schrittwieser, Julian and others},
  journal={arXiv preprint arXiv:2403.05530},
  year={2024}
}

@article{xiong2023effectivelongcontextscalingfoundation,
  title = {Effective Long-Context Scaling of Foundation Models},
  author = {Wenhan Xiong and Jingyu Liu and Igor Molybog and Hejia Zhang and Prajjwal Bhargava and Rui Hou and Louis Martin and Rashi Rungta and Karthik Abinav Sankararaman and Barlas Oguz and Madian Khabsa and Han Fang and Yashar Mehdad and Sharan Narang and Kshitiz Malik and Angela Fan and Shruti Bhosale and Sergey Edunov and Mike Lewis and Sinong Wang and Hao Ma},
  journal = {arXiv preprint arXiv:2309.16039},
  year={2023},
}

@article{hoffmann2022training,
  title={Training compute-optimal large language models},
  author={Hoffmann, Jordan and Borgeaud, Sebastian and Mensch, Arthur and Buchatskaya, Elena and Cai, Trevor and Rutherford, Eliza and Casas, Diego de Las and Hendricks, Lisa Anne and Welbl, Johannes and Clark, Aidan and others},
  journal={arXiv preprint arXiv:2203.15556},
  year={2022}
}

@article{an2024does,
  title={Why Does the Effective Context Length of LLMs Fall Short?},
  author={An, Chenxin and Zhang, Jun and Zhong, Ming and Li, Lei and Gong, Shansan and Luo, Yao and Xu, Jingjing and Kong, Lingpeng},
  journal={arXiv preprint arXiv:2410.18745},
  year={2024}
}

@article{dai2024deepseekmoe,
  title={Deepseekmoe: Towards ultimate expert specialization in mixture-of-experts language models},
  author={Dai, Damai and Deng, Chengqi and Zhao, Chenggang and Xu, RX and Gao, Huazuo and Chen, Deli and Li, Jiashi and Zeng, Wangding and Yu, Xingkai and Wu, Y and others},
  journal={arXiv preprint arXiv:2401.06066},
  year={2024}
}

@article{yang2024qwen2,
  title={Qwen2. 5 Technical Report},
  author={Yang, An and Yang, Baosong and Zhang, Beichen and Hui, Binyuan and Zheng, Bo and Yu, Bowen and Li, Chengyuan and Liu, Dayiheng and Huang, Fei and Wei, Haoran and others},
  journal={arXiv preprint arXiv:2412.15115},
  year={2024}
}

@article{chen2023extendingcontextwindowlarge,
  title={Extending Context Window of Large Language Models via Positional Interpolation}, 
  author={Shouyuan Chen and Sherman Wong and Liangjian Chen and Yuandong Tian},
  year={2023},
  journal={arXiv preprint arXiv:2401.06066},
}

@article{lu2024longheads,
  title={LongHeads: Multi-Head Attention is Secretly a Long Context Processor},
  author={Lu, Yi and Zhou, Xin and He, Wei and Zhao, Jun and Ji, Tao and Gui, Tao and Zhang, Qi and Huang, Xuanjing},
  journal={arXiv preprint arXiv:2402.10685},
  year={2024}
}

@article{zhang2024simlayerkv,
  title={Simlayerkv: A simple framework for layer-level KV cache reduction},
  author={Zhang, Xuan and Du, Cunxiao and Du, Chao and Pang, Tianyu and Gao, Wei and Lin, Min},
  journal={arXiv preprint arXiv:2410.13846},
  year={2024}
}

@article{team2025kimi,
  title={Kimi k1. 5: Scaling Reinforcement Learning with LLMs},
  author={Team, Kimi and Du, Angang and Gao, Bofei and Xing, Bowei and Jiang, Changjiu and Chen, Cheng and Li, Cheng and Xiao, Chenjun and Du, Chenzhuang and Liao, Chonghua and others},
  journal={arXiv preprint arXiv:2501.12599},
  year={2025}
}

@article{guo2025deepseek,
  title={Deepseek-r1: Incentivizing reasoning capability in llms via reinforcement learning},
  author={Guo, Daya and Yang, Dejian and Zhang, Haowei and Song, Junxiao and Zhang, Ruoyu and Xu, Runxin and Zhu, Qihao and Ma, Shirong and Wang, Peiyi and Bi, Xiao and others},
  journal={arXiv preprint arXiv:2501.12948},
  year={2025}
}

@article{guan2024deliberative,
  title={Deliberative alignment: Reasoning enables safer language models},
  author={Guan, Melody Y and Joglekar, Manas and Wallace, Eric and Jain, Saachi and Barak, Boaz and Heylar, Alec and Dias, Rachel and Vallone, Andrea and Ren, Hongyu and Wei, Jason and others},
  journal={arXiv preprint arXiv:2412.16339},
  year={2024}
}

@article{mercat2024linearizing,
  title={Linearizing Large Language Models},
  author={Mercat, Jean and Vasiljevic, Igor and Keh, Sedrick and Arora, Kushal and Dave, Achal and Gaidon, Adrien and Kollar, Thomas},
  journal={arXiv preprint arXiv:2405.06640},
  year={2024}
}

@article{bick2025transformers,
  title={Transformers to ssms: Distilling quadratic knowledge to subquadratic models},
  author={Bick, Aviv and Li, Kevin and Xing, Eric and Kolter, J Zico and Gu, Albert},
  journal={Advances in Neural Information Processing Systems},
  volume={37},
  pages={31788--31812},
  year={2025}
}

@article{wang2024mamba,
  title={The mamba in the llama: Distilling and accelerating hybrid models},
  author={Wang, Junxiong and Paliotta, Daniele and May, Avner and Rush, Alexander M and Dao, Tri},
  journal={arXiv preprint arXiv:2408.15237},
  year={2024}
}

@article{zhang2024lolcats,
  title={LoLCATs: On Low-Rank Linearizing of Large Language Models},
  author={Zhang, Michael and Arora, Simran and Chalamala, Rahul and Wu, Alan and Spector, Benjamin and Singhal, Aaryan and Ramesh, Krithik and R{\'e}, Christopher},
  journal={arXiv preprint arXiv:2410.10254},
  year={2024}
}

@article{sun2023retentive,
  title={Retentive network: A successor to transformer for large language models},
  author={Sun, Yutao and Dong, Li and Huang, Shaohan and Ma, Shuming and Xia, Yuqing and Xue, Jilong and Wang, Jianyong and Wei, Furu},
  journal={arXiv preprint arXiv:2307.08621},
  year={2023}
}

@article{dao2024transformers,
  title={Transformers are SSMs: Generalized models and efficient algorithms through structured state space duality},
  author={Dao, Tri and Gu, Albert},
  journal={arXiv preprint arXiv:2405.21060},
  year={2024}
}

@article{peng2024eagle,
  title={Eagle and finch: Rwkv with matrix-valued states and dynamic recurrence},
  author={Peng, Bo and Goldstein, Daniel and Anthony, Quentin and Albalak, Alon and Alcaide, Eric and Biderman, Stella and Cheah, Eugene and Du, Xingjian and Ferdinan, Teddy and Hou, Haowen and others},
  journal={arXiv preprint arXiv:2404.05892},
  year={2024}
}

@article{li2025minimax,
  title={Minimax-01: Scaling foundation models with lightning attention},
  author={Li, Aonian and Gong, Bangwei and Yang, Bo and Shan, Boji and Liu, Chang and Zhu, Cheng and Zhang, Chunhao and Guo, Congchao and Chen, Da and Li, Dong and others},
  journal={arXiv preprint arXiv:2501.08313},
  year={2025}
}

@article{liu2024retrievalattention,
  title={Retrievalattention: Accelerating long-context llm inference via vector retrieval},
  author={Liu, Di and Chen, Meng and Lu, Baotong and Jiang, Huiqiang and Han, Zhenhua and Zhang, Qianxi and Chen, Qi and Zhang, Chengruidong and Ding, Bailu and Zhang, Kai and others},
  journal={arXiv preprint arXiv:2409.10516},
  year={2024}
}

@article{shazeer2017outrageously,
  title={Outrageously large neural networks: The sparsely-gated mixture-of-experts layer},
  author={Shazeer, Noam and Mirhoseini, Azalia and Maziarz, Krzysztof and Davis, Andy and Le, Quoc and Hinton, Geoffrey and Dean, Jeff},
  journal={arXiv preprint arXiv:1701.06538},
  year={2017}
}

@article{lepikhin2020gshard,
  title={Gshard: Scaling giant models with conditional computation and automatic sharding},
  author={Lepikhin, Dmitry and Lee, HyoukJoong and Xu, Yuanzhong and Chen, Dehao and Firat, Orhan and Huang, Yanping and Krikun, Maxim and Shazeer, Noam and Chen, Zhifeng},
  journal={arXiv preprint arXiv:2006.16668},
  year={2020}
}

@article{fedus2022switch,
  title={Switch transformers: Scaling to trillion parameter models with simple and efficient sparsity},
  author={Fedus, William and Zoph, Barret and Shazeer, Noam},
  journal={Journal of Machine Learning Research},
  volume={23},
  number={120},
  pages={1--39},
  year={2022}
}

@article{zoph2022st,
  title={St-moe: Designing stable and transferable sparse expert models},
  author={Zoph, Barret and Bello, Irwan and Kumar, Sameer and Du, Nan and Huang, Yanping and Dean, Jeff and Shazeer, Noam and Fedus, William},
  journal={arXiv preprint arXiv:2202.08906},
  year={2022}
}

@article{milakov2018onlinenormalizercalculationsoftmax,
      title={Online normalizer calculation for softmax}, 
      author={Maxim Milakov and Natalia Gimelshein},
      year={2018},
      journal={arXiv preprint arXiv:1805.02867},
}

@article{liu2023blockwiseparalleltransformerlarge,
      title={Blockwise Parallel Transformer for Large Context Models}, 
      author={Hao Liu and Pieter Abbeel},
      year={2023},
      journal={arXiv preprint arXiv:2305.19370},
}

@inproceedings{waswani2017attention,
  title={Attention is all you need},
  author={Waswani, A and Shazeer, N and Parmar, N and Uszkoreit, J and Jones, L and Gomez, A and Kaiser, L and Polosukhin, I},
  booktitle={NIPS},
  year={2017}
}

@misc{claude,
title = {Introducing 100K Context Windows},
    author={Anthropic},
    howpublished={\url{https://www.anthropic.com/news/100k-context-windows}},
    year={2023},
}

@misc{kimi,
    title={Kimi Chat},
    author={MoonshotAI},
    howpublished={\url{https://kimi.moonshot.cn/}},
    year={2023},
}

@inproceedings{rajbhandari2022deepspeed,
  title={Deepspeed-moe: Advancing mixture-of-experts inference and training to power next-generation ai scale},
  author={Rajbhandari, Samyam and Li, Conglong and Yao, Zhewei and Zhang, Minjia and Aminabadi, Reza Yazdani and Awan, Ammar Ahmad and Rasley, Jeff and He, Yuxiong},
  booktitle={International conference on machine learning},
  pages={18332--18346},
  year={2022},
  organization={PMLR}
}

@article{dao2022flashattention,
  title={Flashattention: Fast and memory-efficient exact attention with io-awareness},
  author={Dao, Tri and Fu, Dan and Ermon, Stefano and Rudra, Atri and R{\'e}, Christopher},
  journal={Advances in Neural Information Processing Systems},
  volume={35},
  pages={16344--16359},
  year={2022}
}

@article{shoeybi2019megatron,
  title={Megatron-lm: Training multi-billion parameter language models using model parallelism},
  author={Shoeybi, Mohammad and Patwary, Mostofa and Puri, Raul and LeGresley, Patrick and Casper, Jared and Catanzaro, Bryan},
  journal={arXiv preprint arXiv:1909.08053},
  year={2019}
}

@article{watson2025human,
  title={Human hippocampal CA3 uses specific functional connectivity rules for efficient associative memory},
  author={Watson, Jake F and Vargas-Barroso, Victor and Morse-Mora, Rebecca J and Navas-Olive, Andrea and Tavakoli, Mojtaba R and Danzl, Johann G and Tomschik, Matthias and R{\"o}ssler, Karl and Jonas, Peter},
  journal={Cell},
  volume={188},
  number={2},
  pages={501--514},
  year={2025},
  publisher={Elsevier}
}

\appendix
\section{Appendix}

\subsection{Long Context Scalability}\label{subsec:trailing_scaling}
To address the bias in natural data distribution that favors short contexts, we strategically segmented the overall sequences into discrete segments based on their actual positions. For example, the segment spanning positions 30K-32K exclusively reflects losses associated with documents exceeding 30K context lengths and also masks the positions from 30K to 32K. This approach ensures a more balanced and representative evaluation across different context lengths.
In our exploration of long-context scalability, we made a pivotal discovery: the trailing tokens account for the majority of the performance discrepancy between the full context baseline and the newly proposed sparse attention architectures. Consequently, we streamlined the long-context scaling process by focusing on trailing token scaling. This not only simplifies the computational requirements but also significantly enhances the efficiency and effectiveness of investigating long-context scenarios. This finding holds substantial implications for the development of more efficient and scalable attention mechanisms in the future.

\begin{figure}[htbp]
  \centering
  \subfloat[Scaling law (0-2k) \label{fig1}]{\includegraphics[width=0.22\textwidth]{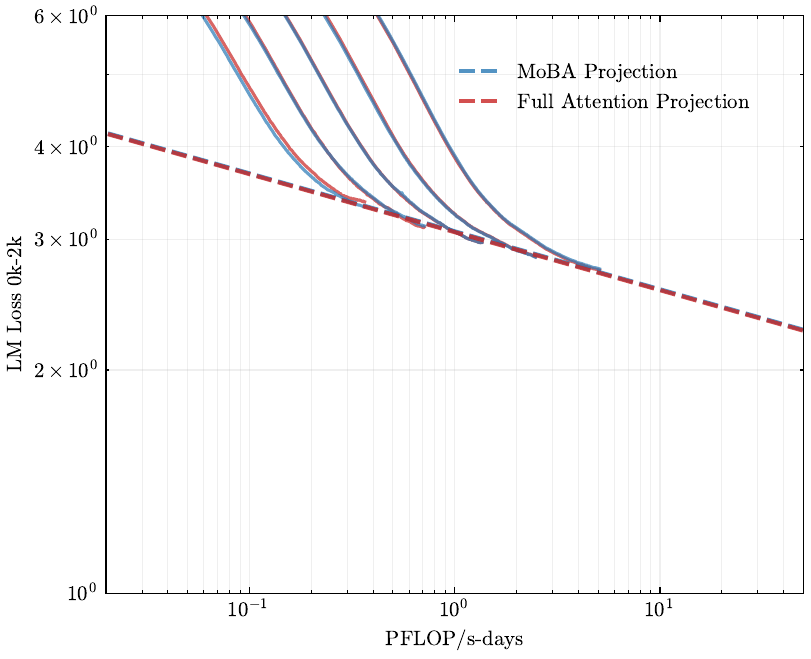}}\hfill
  \subfloat[Scaling law (2-4k) \label{fig2}]{\includegraphics[width=0.22\textwidth]{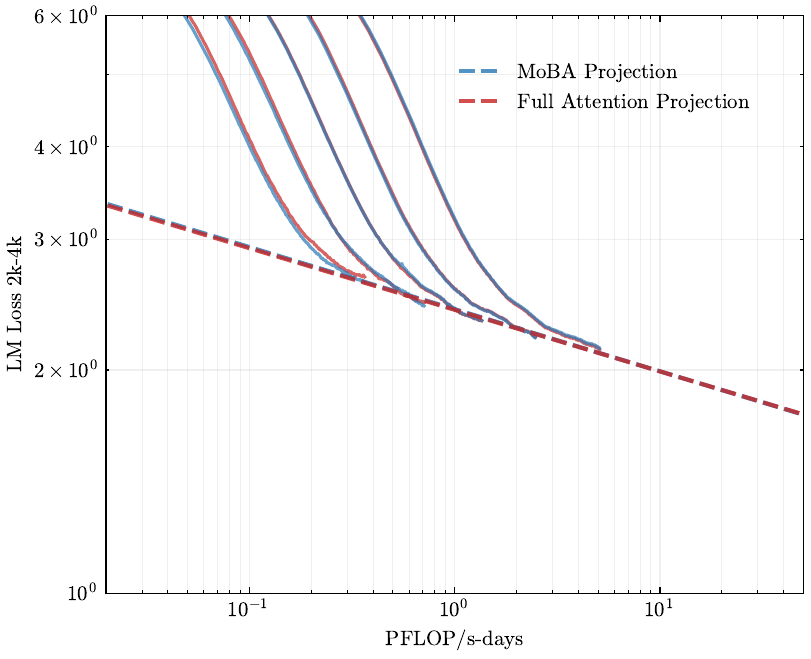}}\hfill
  \subfloat[Scaling law (4-6k) \label{fig3}]{\includegraphics[width=0.22\textwidth]{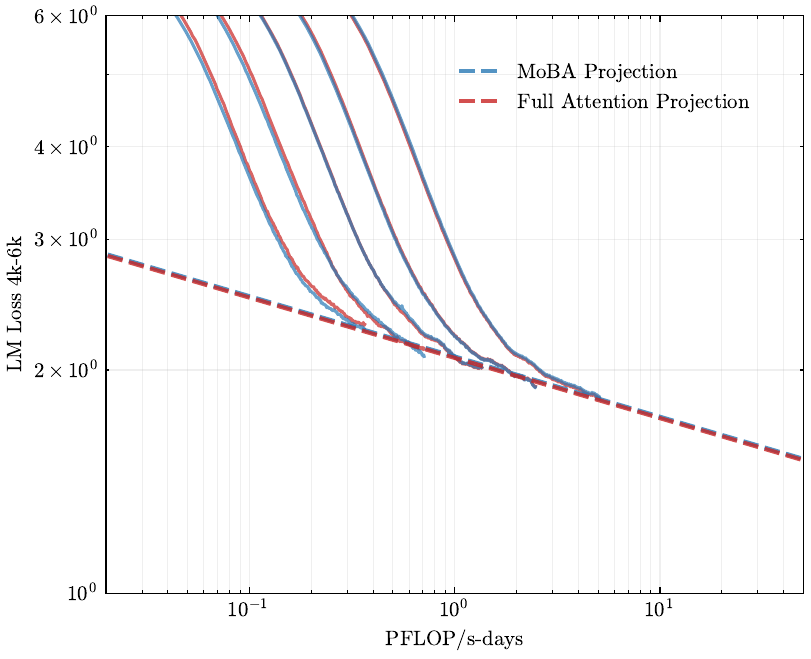}}\hfill
  \subfloat[Scaling law (6-8k) \label{fig4}]{\includegraphics[width=0.22\textwidth]{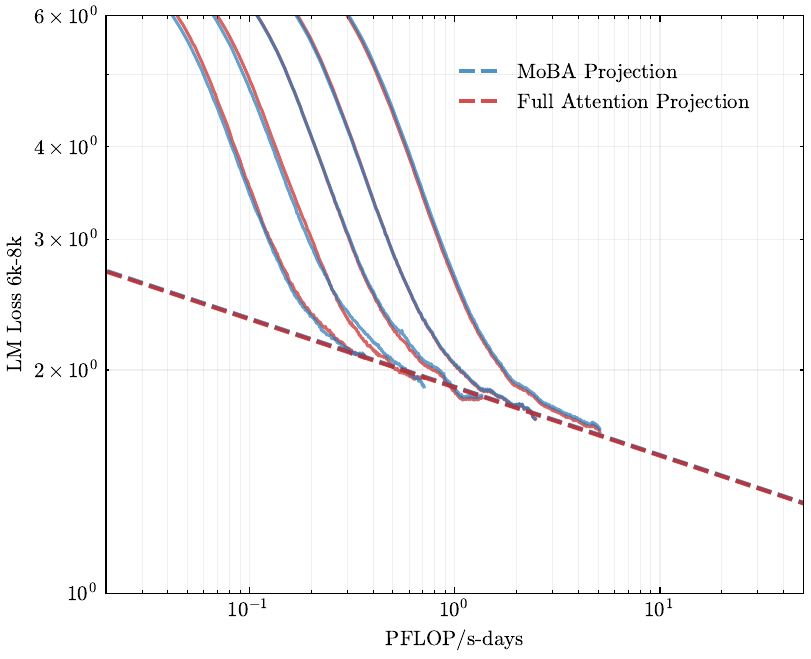}}\\
  \subfloat[Scaling law (8-10k) \label{fig5}]{\includegraphics[width=0.22\textwidth]{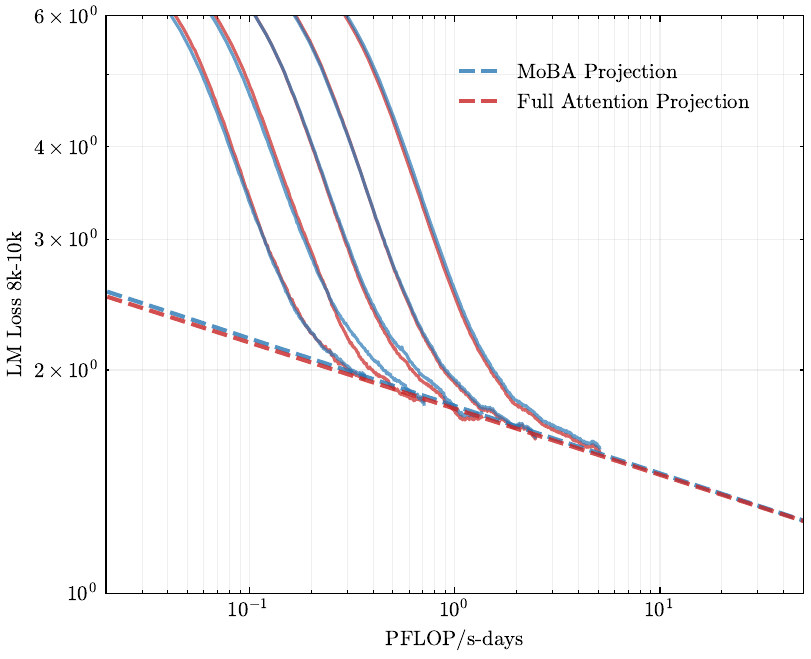}}\hfill
  \subfloat[Scaling law (10-12k) \label{fig6}]{\includegraphics[width=0.22\textwidth]{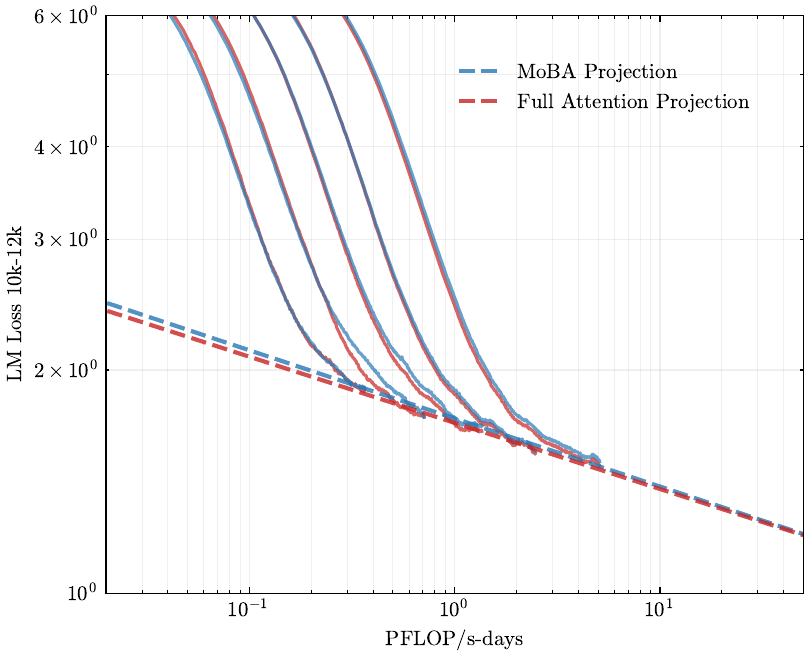}}\hfill
  \subfloat[Scaling law (12-14k) \label{fig7}]{\includegraphics[width=0.22\textwidth]{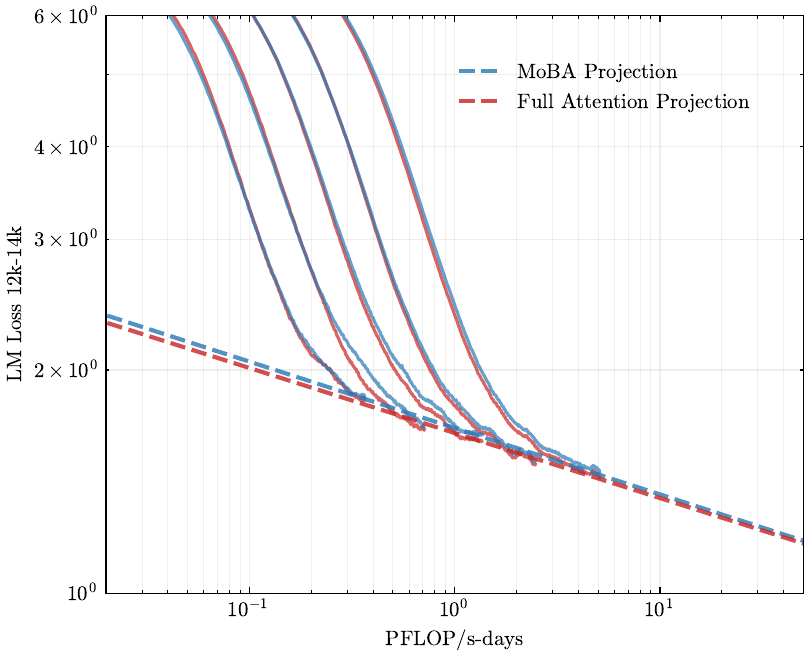}}\hfill
  \subfloat[Scaling law (14-16k) \label{fig8}]{\includegraphics[width=0.22\textwidth]{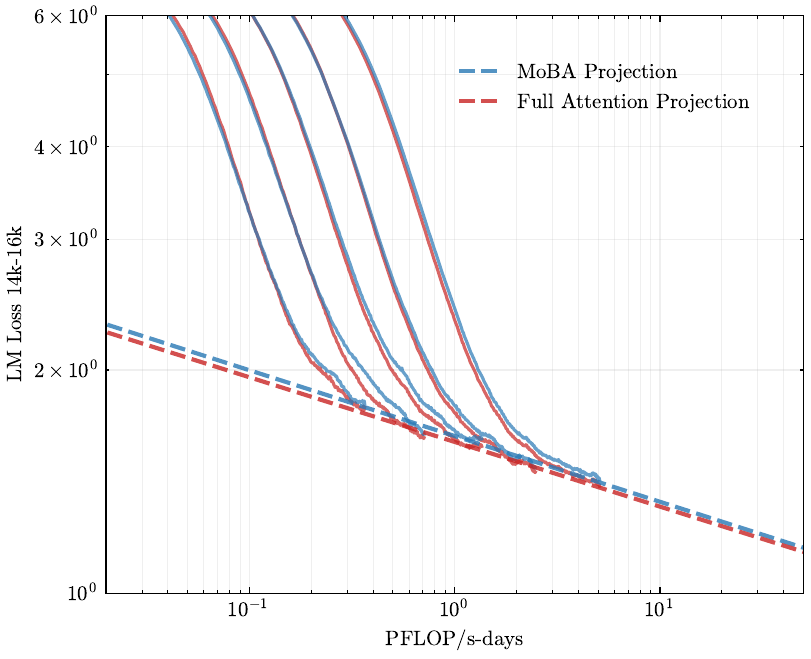}}
  \caption{Scaling laws for positions 0-16k}
  \label{fig:page1}
\end{figure}

\begin{figure}[htbp]
  \ContinuedFloat
  \centering
  \subfloat[Scaling law (16-18k) \label{fig9}]{\includegraphics[width=0.22\textwidth]{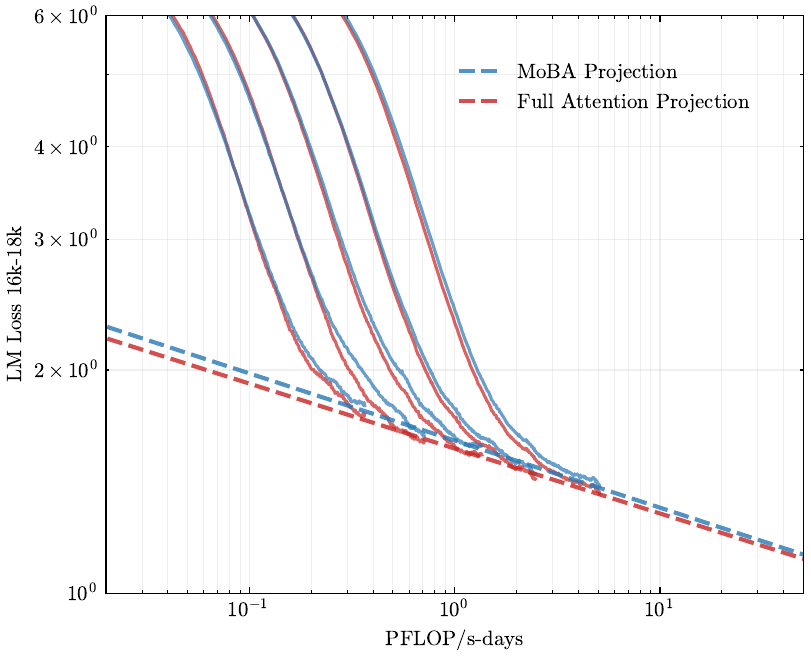}}\hfill
  \subfloat[Scaling law (18-20k) \label{fig10}]{\includegraphics[width=0.22\textwidth]{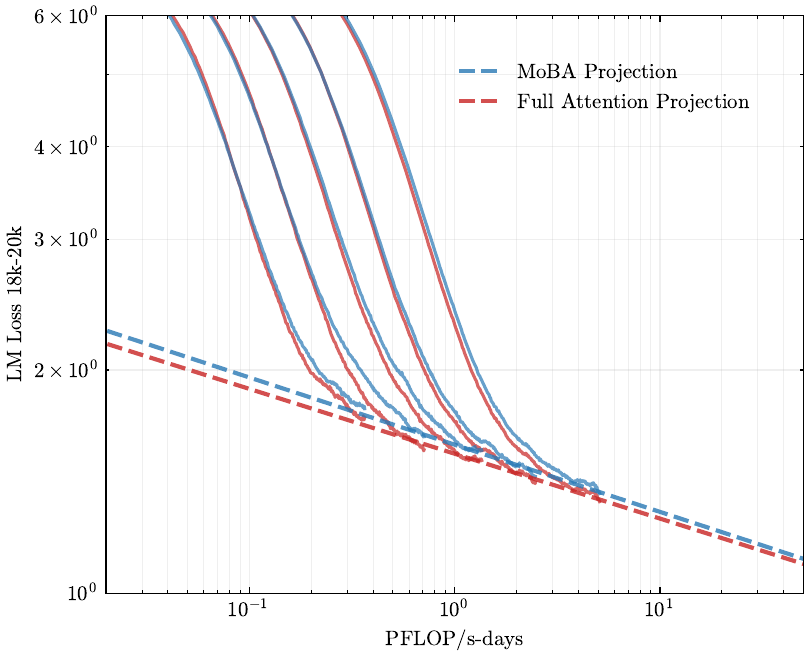}}\hfill
  \subfloat[Scaling law (20-22k) \label{fig11}]{\includegraphics[width=0.22\textwidth]{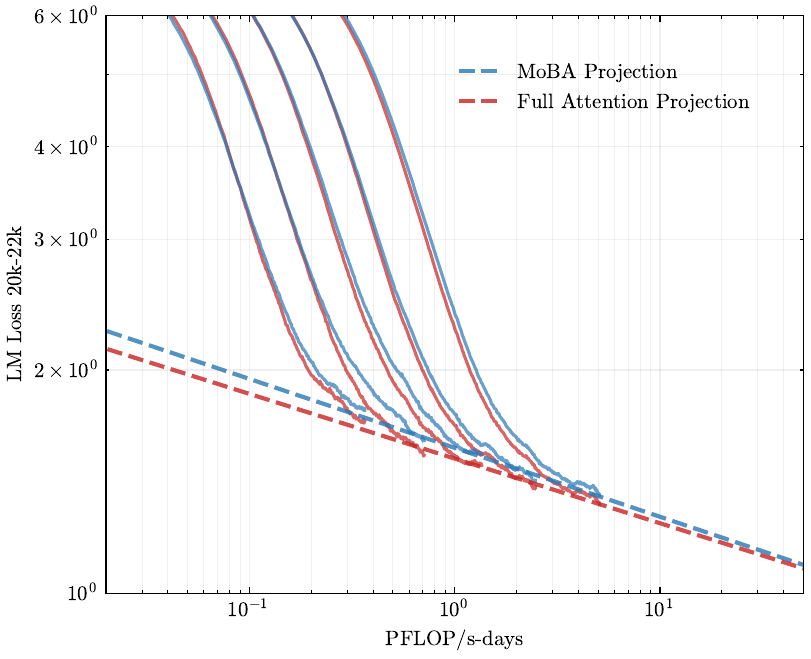}}\hfill
  \subfloat[Scaling law (22-24k) \label{fig12}]{\includegraphics[width=0.22\textwidth]{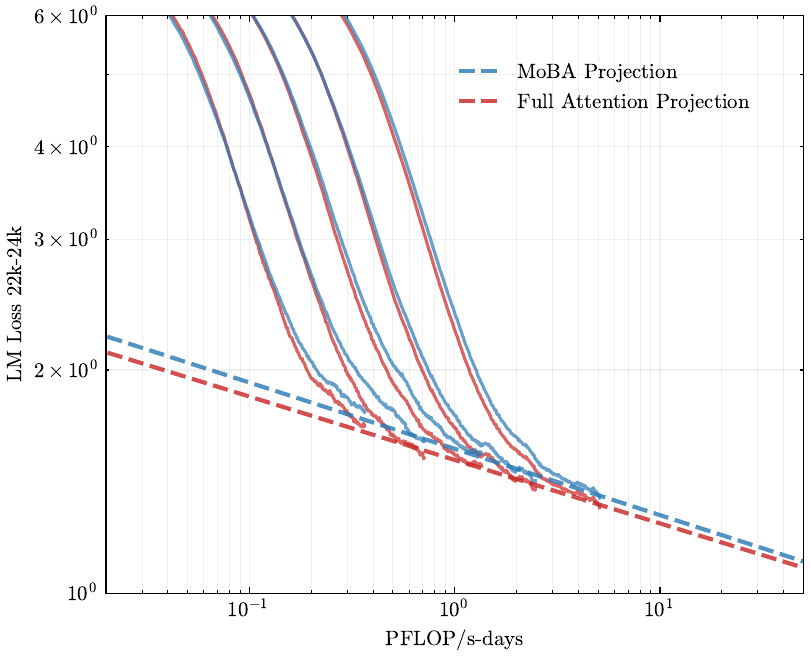}}\\
  \subfloat[Scaling law (24-26k) \label{fig13}]{\includegraphics[width=0.22\textwidth]{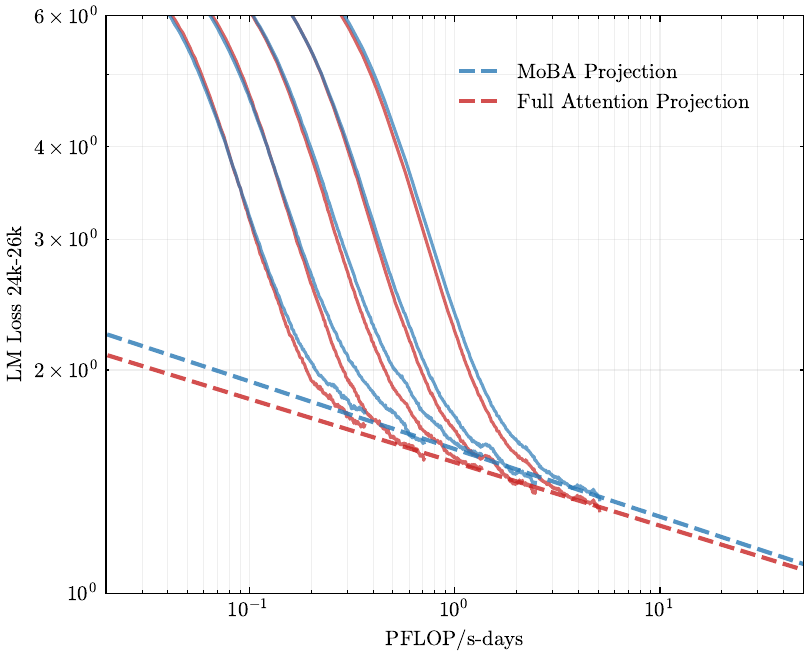}}\hfill
  \subfloat[Scaling law (26-28k) \label{fig14}]{\includegraphics[width=0.22\textwidth]{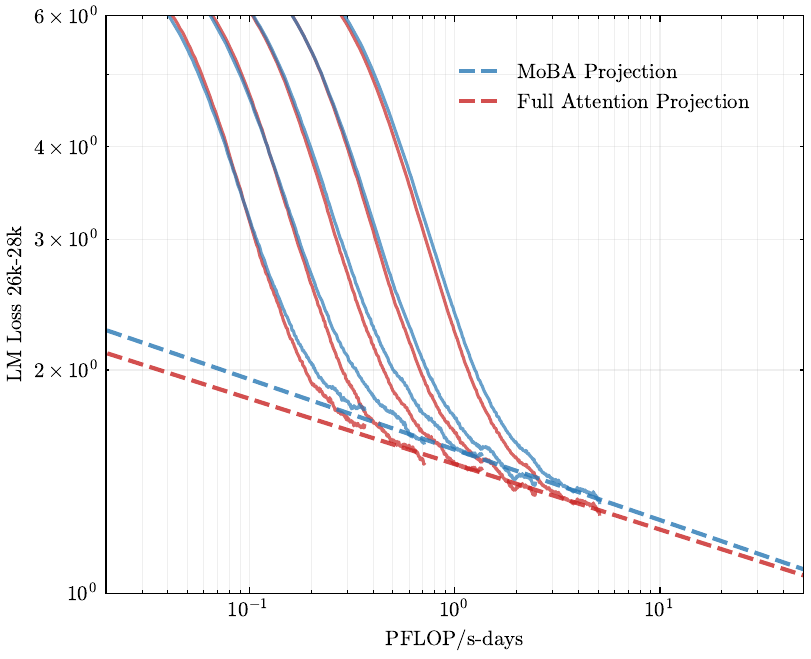}}\hfill
  \subfloat[Scaling law (28-30k) \label{fig15}]{\includegraphics[width=0.22\textwidth]{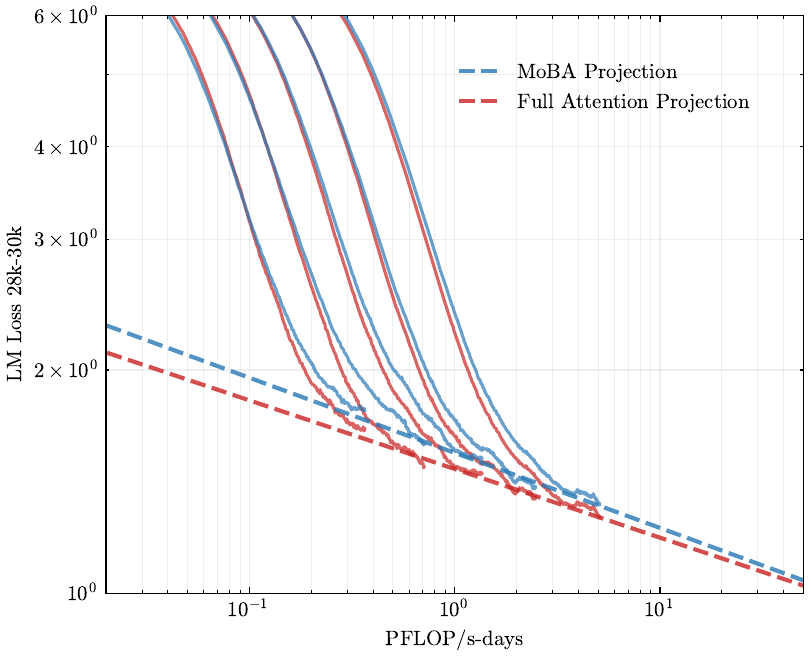}}\hfill
  \subfloat[Scaling law (30-32k) \label{fig16}]{\includegraphics[width=0.22\textwidth]{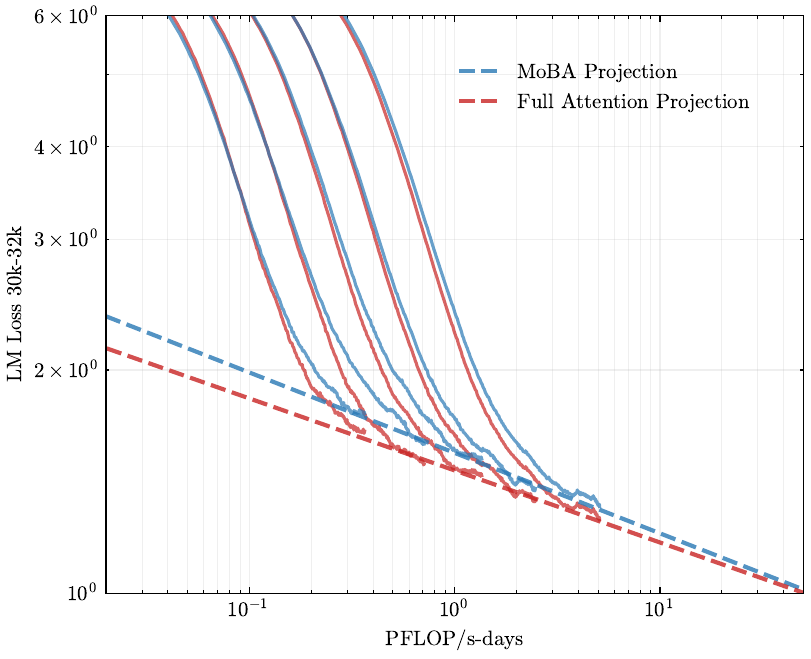}}
  \caption{Scaling laws for positions 16-32k}
  \label{fig:page2}
\end{figure}

\begin{table}
\caption{Loss scaling with different positions}
\small
\begin{center}
\begin{tabular}{l|c|c}
\toprule
LM Loss Position Range & MoBA & Full \\
\midrule
0K - 2K & $3.075 \times C^{-0.078}$ & $3.068 \times C^{-0.078}$ \\
2K - 4K & $2.415 \times C^{-0.084}$ & $2.411 \times C^{-0.083}$ \\
4K - 6K & $2.085 \times C^{-0.081}$ & $2.077 \times C^{-0.081}$ \\
6K - 8K & $1.899 \times C^{-0.092}$ & $1.894 \times C^{-0.092}$ \\
8K - 10K & $1.789 \times C^{-0.091}$ & $1.774 \times C^{-0.089}$ \\
10K - 12K & $1.721 \times C^{-0.092}$ & $1.697 \times C^{-0.087}$ \\
12K - 14K & $1.670 \times C^{-0.089}$ & $1.645 \times C^{-0.088}$ \\
14K - 16K & $1.630 \times C^{-0.089}$ & $1.600 \times C^{-0.087}$ \\
16K - 18K & $1.607 \times C^{-0.090}$ & $1.567 \times C^{-0.087}$ \\
18K - 20K & $1.586 \times C^{-0.091}$ & $1.542 \times C^{-0.087}$ \\
20K - 22K & $1.571 \times C^{-0.093}$ & $1.519 \times C^{-0.086}$ \\
22K - 24K & $1.566 \times C^{-0.089}$ & $1.513 \times C^{-0.085}$ \\
24K - 26K & $1.565 \times C^{-0.091}$ & $1.502 \times C^{-0.085}$ \\
26K - 28K & $1.562 \times C^{-0.095}$ & $1.493 \times C^{-0.088}$ \\
28K - 30K & $1.547 \times C^{-0.097}$ & $1.471 \times C^{-0.091}$ \\
30K - 32K & $1.546 \times C^{-0.108}$ & $1.464 \times C^{-0.097}$ \\
\bottomrule
\end{tabular}
\end{center}
\end{table}

\end{document}